\documentclass[lettersize,journal]{IEEEtran}
\usepackage{mathtools,amsmath,amsfonts,amsthm,amssymb}
\usepackage{algorithmic,algorithm}
\usepackage{array}
\usepackage[caption=false,font=normalsize,labelfont=sf,textfont=sf]{subfig}
\usepackage{textcomp}
\usepackage{stfloats}
\usepackage{url}
\usepackage{bm}
\usepackage{verbatim}
\usepackage{graphicx}
\usepackage{lipsum}
\usepackage{epstopdf}
\usepackage{enumitem}
\usepackage{parskip} 
\usepackage[pdfencoding=auto]{hyperref}
\usepackage{cleveref}
\usepackage{xcolor}

\def\IR {\mathbb{R}}  % Real set
\def\IZ {\mathbb{Z}}  % integer set
\def\IRn{\IR^{n}   }  % Rn 
\def\cP {\mathcal{P} }  
\def\cL {\mathcal{L} }
\DeclareMathOperator*{\argmin}{\textrm{argmin}}  % argmin
\DeclareMathOperator*{\argmax}{\textrm{argmax}}  % argmax
\newtheorem{theorem}{Theorem}
\newtheorem{lemma}[theorem]{Lemma}

\newlength\myindent
\newcommand\bindent{%
  \begingroup
  \setlength{\itemindent}{\myindent}
  \addtolength{\algorithmicindent}{\myindent}
}
\newcommand\eindent{\endgroup}

\DeclareMathOperator{\Tr}{\textrm{tr}}

\ifpdf
  \DeclareGraphicsExtensions{.eps,.pdf,.png,.jpg}
\else
  \DeclareGraphicsExtensions{.eps}
\fi

\def\BibTeX{{\rm B\kern-.05em{\sc i\kern-.025em b}\kern-.08em
    T\kern-.1667em\lower.7ex\hbox{E}\kern-.125emX}}
\usepackage{balance}
\begin{document}
\title{Inhomogeneous graph trend filtering via a $\ell_{2,0}$-norm cardinality penalty}
\author{IEEE Publication Technology Department
\thanks{Manuscript created October, 2020; This work was developed by the IEEE Publication Technology Department. This work is distributed under the \LaTeX \ Project Public License (LPPL) ( http://www.latex-project.org/ ) version 1.3. A copy of the LPPL, version 1.3, is included in the base \LaTeX \ documentation of all distributions of \LaTeX \ released 2003/12/01 or later. The opinions expressed here are entirely that of the author. No warranty is expressed or implied. User assumes all risk.}}
\author{Xiaoqing Huang$^*$\thanks{X. Huang and K. Huang are with the Dept. of Biostatistics and Health Data Science, Indiana University School of Medicine, Indianapolis, IN 46202, USA.}, Andersen Ang$^*$\thanks{A. Ang is with the School of Electronics and Computer Science, University of Southampton, Southampton, SO17 1BJ, UK.}, Kun Huang\thanks{J. Zhang is with the Dept. of Medical and Molecular Genetics, Indiana University, Indianapolis, IN 46202, USA.
}, Jie Zhang\thanks{Y. Wang is with the Dept. of Computer Science, Indiana University, Bloomington, IN 47408, USA. Email: yijwang@iu.edu}, and Yijie Wang\thanks{This work is supported in part by NIH under grant R35GM147241.}\thanks{$^*$X. Huang and A. Ang are equal contributotrs to this work.}}

\markboth{Journal of \LaTeX\ Class Files,~Vol.~18, No.~9, September~2020}%
{How to Use the IEEEtran \LaTeX \ Templates}

\maketitle

\begin{abstract}
We study estimation of piecewise smooth signals over a graph.
We propose a $\ell_{2,0}$-norm penalized Graph Trend Filtering (GTF) model to estimate piecewise smooth graph signals that exhibit inhomogeneous levels of smoothness across the nodes. 
We prove that the proposed GTF model is simultaneously a k-means clustering on the signal over the nodes and a minimum graph cut on the edges of the graph, where the clustering and the cut share the same assignment matrix.
We propose two methods to solve the proposed GTF model: a spectral decomposition method and a method based on simulated annealing.
In the experiment on synthetic and real-world datasets, we show that the proposed GTF model has a better performances compared with existing approaches on the tasks of denoising, support recovery and semi-supervised classification. 
We also show that the proposed GTF model can be solved more efficiently than existing models for the dataset with a large edge set.
\end{abstract}


\section{Introduction}
\begin{figure*}[t!]
\includegraphics[width=\textwidth]{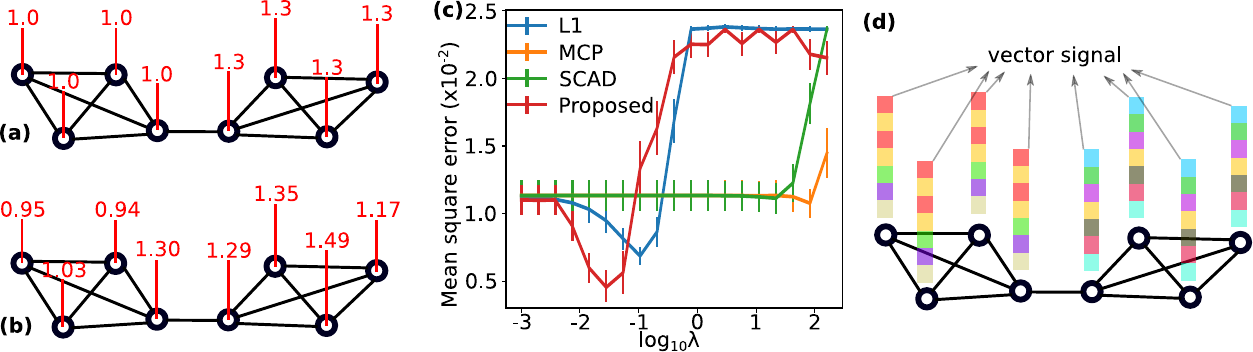}
\caption{\textup{\textbf{Toy examples.}}
(a) The ground truth of an inhomogeneous scalar-valued graph signal.
(b) The same graph signal in (a) with added Gaussian noise $\mathcal{N}(0,0.1)$.
(c) Comparison of the recovery of the ground truth in (a) from 20 different noisy inhomogeneous graph signals as in (b) with different parameter $\lambda$ in~\eqref{prob:0} and~\eqref{l0model}.
%L0 stands for the proposed model in~\eqref{l0model}.
The proposed model~\eqref{l0model} achieves the lowest mean square error when screening the parameter $\lambda$. 
(d) An example of a vector-valued graph signal.}
\label{fig-toy}
\end{figure*}

\IEEEPARstart{E}{stimating} signal from noisy observations is a classical problem in signal processing.
This work focuses on Graph Signal Estimation (GSE), which is a powerful tool for analyzing signals that are defined on irregular and complex domains (e.g., social networks, brain networks, and sensor networks) and has a lot of real-world applications~\cite{7208894,7117446,7605501,4526700}.
Fig.~\ref{fig-toy}a gives an illustration of a graph signal. 
In a graph signal, the signal is residing on each node in a graph. 
The signal could be a scalar-valued signal (as shown in Fig.~\ref{fig-toy}a) or vector-valued signal (as shown Fig.~\ref{fig-toy}c).
The edges between nodes in the graph introduce additional information and assumption for the signals residing on the nodes.
A typical assumption is that the graph signal is globally smooth with respect to the intrinsic structure of the underlying graph, i.e., the signals on the neighbor nodes tend to be similar, or in other words, the local variation of the signals around each node is low. 
Such global smoothness of a graph signal is further mathematically defined and quantitatively measured by the $p$-Dirichlet form~\cite{6494675,8347162,10.1007/978-3-540-27819-1_43,10.1007/978-3-540-72823-8_12,4526700,10.1109/FOCS.2007.66} and the graph Laplacian quadratic form ($p$-Dirichlet form with $p=2$)~\cite{10.1109/FOCS.2007.66}.  
GSE based on such global smoothness assumption has been extensively investigated in graph signal processing~\cite{6494675,8347162,7208894,7117446,7605501,4526700} and in the context of Laplacian regularization~\cite{10.1007/978-3-540-27819-1_43,10.5555/3041838.3041953}.

Although many existing works~\cite{6494675,8347162,7208894,7117446,7605501,4526700} assume the global smoothness on the graph signal,
many real-world graph signals exhibit inhomogeneous levels of smoothness over the graph~\cite{8926407,pmlr-v38-wang15d,pmid21527005}. In other words, there is a discontinuity in the first-order behaviour of the signal value across the graph, see Fig.~\ref{fig-toy}(a) for a toy illustration. As shown the signals within the communities are the same (homogeneous) but the signals between two communites are different (inhomogeneous).  Such an ``intra-community homogeneous, inter-community inhomogeneous'' property indicates that the graph signals have large variations between different regions and have a small variation within regions (as shown in Fig.~\ref{fig-toy}a). 

\textcolor{black}{
In real-world applications, we have many graph signals that possess the ``inhomogeneous levels of smoothness" property. 
In bioinformatics and computational biology, tremendous efforts have been made to build molecular networks for understanding biological processes, identifying disease mechanisms, predicting phenotypic outcomes, etc. 
These molecular networks include but are not limited to protein-protein interaction networks, gene regulatory networks, metabolic networks, and pathway networks. 
Many databases provide access to these molecular networks~\cite{szklarczyk2023string,Kim2022-fy,Rodchenkov2020-fr}.
For example, the database STRING~\cite{szklarczyk2023string} collects functional protein association networks from literature curation, computational predictions, and model organisms based on orthology. 
The database HumanNet~\cite{Kim2022-fy} provides human gene networks for disease research.
The database Pathway Commons~\cite{Rodchenkov2020-fr} aggregates interactions from pathway and interaction databases. 
For all these molecular networks, when we add a patient-level or cell-level data (e.g. gene expression, mutation, copy numbers, etc) on top of them, we have generated graph signals that exhibit some levels of inhomogeneousity of smoothness across the graph.  
For such graph signals, from the system biology point of view~\cite{Hofree2013-ot}, a valid assumption is that the signal (e.g., gene expression, mutation, etc) within a graph community tends to be homogeneous (i.e., having similar values or having a smooth variation across the nodes in the same community), but such signal across the different communities can be highly different (i.e., inhomogeneous). 
For example, in gene co-expression networks~\cite{pmid12934013}, gene expressions within a gene community tend to have similar expression values, but the gene expression across the communities is highly different. 
GSE on the graph signal of these molecular networks can help disease stratification~\cite{Hofree2013-ot} through denoising the patient-level or the cell-level measurements we put on top of the molecular networks. {Another example is the graph signal on the protein-protein interaction network. Here, nodes are genes, and edges are protein-protein interactions measured by biological experiments. In such kind of network, for each node (gene), we could further experimentally measure its expression, and thus, we can establish a graph and the signal on the graph. To fulfill certain cellular functions, the genes do not work alone but work collectively, as illustrated in a graph. As such, the expression of the genes should be similar within the same functional groups but different across functional groups, which is exactly the property of the inhomogeneous graph signal. Additionally, the inhomogeneous graph signal could manifest on MIR images of the brain. Using several MRI scans of the brain, we can construct a consensus brain network, where nodes are voxels and edges are correlations between measurements on the voxels. For a given MRI, we have a consensus brain network and measurement on each voxel of the network. Based on the hypothesis of neural synchrony and neural oscillations in neuroscience, brain signals over the graph within the same functional region will exhibit the property of inhomogeneous graph signals.  }
} 

\textcolor{black}{
The existence of real-world piecewise smooth graph signals (especially in computational biology illustrated in the previous paragaph) motivates us to study GSE for such graph signals.
}
Therefore, in this paper, we propose a model: a $\ell_{2,0}$-norm penalized \textbf{1st-order Graph Trend Filtering (GTF)}~\cite{pmlr-v38-wang15d}, for denoising and extracting patterns of the signal on a graph with localized discontinuities.
Before we state explicitly our contribution, below we first review about the GTF model.

\subsection{GTF for scalar-valued graph signals}
Given a graph $G(V,E)$ of $n$ nodes $V=\{1,\dots, n\}$ and edges $E=\{(i,j) \,|\, i,j \in V \text{ that are connected}\}$, let $\bm{y}  \in \IRn$ be the observed signal measurement over the nodes, i.e.,  the element $y_i\in \IR$ is the scalar value at node $i$. 
The goal of the 1st-order GTF for scalar-valued graph signals is to find the estimated graph signal $\bm{\beta} \in \IR^n$, where the $i$th element in $\bm{\beta}$ denoted by $\beta_i$ (the estimated signal at node $i$), 
by solving
\begin{equation}\label{prob:0}
\argmin_{\bm{\beta}\in \IRn} 
\underbrace{\dfrac{1}{2}
\|
\bm{y}-\bm{\beta}
\|_2^2}_{\text{fitting / denoise}}
+ 
\lambda 
\underbrace{\sum_{(i,j)\in E} \hspace{-1mm} \Omega (\beta_i, \beta_j)}_{\text{GTF regularization}}
.
\end{equation}
In \eqref{prob:0}, $\lambda \geq 0$ is a regularization parameter and $\Omega(\beta_i,  \beta_j) : \IR \times \IR \rightarrow \IR$ is a penalty term that penalizes the difference between the signals on nodes $\{i,j\}$. 
In other words, $\Omega(\beta_i, \beta_j)$ is a penalization of the neighborhood discrepancy of the graph signal, and it encourages the graph signal value at node $\{i,j\}$ to be similar. 
Different $\Omega$ have been proposed in the literature for GTF, see Table~\ref{tab:Omega}.

\renewcommand{\arraystretch}{2.75}
\begin{table}[]
\caption{Different \,$\Omega$\, proposed in the literature for GTF.}
\centering\small
\begin{tabular}{c||ccc}
Name                   & $\Omega(\beta_i,  \beta_j)$ & Param.  \\  \hline\hline
$\ell_1$-norm~\cite{pmlr-v38-wang15d} & $|\beta_i - \beta_j|$
 & -
\\
SCAD
~\cite{8926407} & $\displaystyle \int_0^{|\beta_i-\beta_j|} \min\left ( 1, \frac{\left ( \gamma-u/\lambda \right )_+}{\gamma-1} \right ) du$   & $\gamma \geq 2$
\\
MCP~\cite{8926407} & $\displaystyle \int_0^{|\beta_i-\beta_j|} \left ( 1-\frac{u}{\gamma \lambda} \right )_+ du$ & $\gamma$
\end{tabular}
\label{tab:Omega}
\end{table}

\subsubsection{GTF with the counting norm penalty}
In this paper, we propose the norm $\Omega$ as $\Omega(\beta_i,\beta_j)
\,=\,
\mathbb{I}_{\neq 0}(\beta_i - \beta_j)$, and propose the GTF model as 
\begin{equation}\label{l0model}
\argmin_{\bm{\beta}\in \IRn} 
\frac{1}{2}  \|\bm{y}-\bm{\beta} \|_2^2
+ 
\lambda 
\sum_{(i,j)\in E} \mathbb{I}_{\neq 0}(\beta_i-\beta_j)
,
\end{equation}
where $\mathbb{I}_{\neq 0} : \IR \rightarrow \IR$ is an indicator function defined as $\mathbb{I}_{\neq 0}(x) = 0$ if $x = 0$ and $\mathbb{I}_{\neq 0}(x) = 1$ otherwise.
In summary, we propose to use a counting norm for $\Omega$, which counts the number of edges that connects nodes with different signal value.

\subsubsection{The advantage of using the proposed norm penalty}\label{advantage} 
The current 1st-order GTF model~\eqref{prob:0} uses  $\ell_1$-norm~\cite{pmlr-v38-wang15d}, Smoothly Clipped Absolute Deviation (SCAD)~\cite{8926407}, or Minimax Concave Penalty (MCP)~\cite{8926407} to penalize the difference between the signals over connected nodes (as shown in Table~\ref{tab:Omega}). 
However, the resulting estimates generated by these penalties are ``biased toward zero'': the $\ell_1$ penalty enforces the minimization of the term $|\beta_i - \beta_j|$ and, therefore, impose $|\beta_i - \beta_j|$ shrink to zero. As a result, the $\ell_1$ penalty forces $\beta_i$ and $\beta_j$ to be numerically close even if they are not. While the proposed counting norm penalty applies the same penalty to $|\beta_i - \beta_j|$ no matter how different $\beta_i$ and $\beta_j$ are. Therefore,  $|\beta_i - \beta_j|$ is not enforced to shrink to zero when $\beta_i$ and $\beta_j$ are different. The same comment also applied to SCAD and MCP penalties.  

We further illustrate such advantage of the proposed GTF model~\eqref{l0model} numerically in Fig.~\ref{fig-toy}c: after screening over the model hyper-parameter ($\lambda$ in~\eqref{prob:0} and~\eqref{l0model}, respectively), the GTF model~\eqref{l0model} achieves the lowest mean square error to recover the ground-truth signal as shown in Fig.~\ref{fig-toy}(a) from the noisy signal as shown in Fig.~\ref{fig-toy}(b).

\subsection{GTF for vector-valued graph signals}
The previous discussion focused on scalar-valued graph signals.
Now we extend the model~\eqref{l0model} to vector-valued graph signals (See Fig.~\ref{fig-toy}d for a pictorial illustration): let $\bm{y}_i\in \IR^d$ be the vector-valued signal at node $i$, let $\bm{Y}=[\bm{y}_1, \dots, \bm{y}_n]^\top  \in \IR^{n\times d}$ be the data matrix (the rows are the observed vector-valued signals on nodes), we consider the optimization problem
\begin{equation}\label{prob:1}
\cP_0 ~:~
 \argmin_{\bm{B} \in \IR^{n \times d}}
\frac{1}{2}\|\bm{Y}-\bm{B}\|_F^2
+
\lambda \hspace{-2mm}
\underbrace{ \sum_{(i,j)\in E}   \hspace{-1mm}
\mathbb{I}_{\neq 0} \Big( \|\bm{b}_i-\bm{b}_j  \|_2 \Big)}_{
\eqqcolon \cL(\bm{B})},
\end{equation}
where $\bm{B}=[\bm{b}_1, ..., \bm{b}_n]^\top \in\IR^{n\times d}$ and $\bm{b}_i \in \IR^d$ is the GTF signal estimate at node $i$. 
Note that~\eqref{l0model} is equivalent to~\eqref{prob:1} when the dimensionality $d=1$.

\paragraph*{On the term \texorpdfstring{$\mathcal{L}(\bm{B})$}{}}
We call $\mathcal{L}(\bm{B})$ in~\eqref{prob:1} the ``$\ell_{2,0}$-norm'' in this paper.
It can be treated as a ``group counting norm'', which is the limiting case of the matrix $\ell_{p,q}$-norm with $p=2$ and $q \rightarrow 0$ on a matrix $\bm{M}$ whose columns are the pairwise difference $\bm{b}_i - \bm{b}_j $  of $\bm{M}$ defined by the edge set $E$, i.e.,
\[
\mathcal{L}(\bm{B}) 
~=~
\lim_{q \rightarrow 0} \| \bm{M} \|_{2,q} 
~\coloneqq~
\lim_{q \rightarrow 0} 
\Bigg(
\sum_{j=1}^{  |E|  } 
\bigg( 
\sqrt{
\sum_{i=1}^d M_{ij}^2
}
\bigg)^q
\Bigg)^{\frac{1}{q}}.
\]
If $\mathcal{L}(\bm{B}) = 0$ in $\cP_0$, it means that all the node signal estimates $\bm{b}_i$ share the same value, or equivalently each column of $\bm{B}$ is a constant vector.

\subsection{Contributions and paper organization} 
Grounded on the advantage of using $\mathcal{L}$ (see \cref{advantage}) that it overcomes the problems of using $\ell_1$-norm~\cite{pmlr-v38-wang15d}, SCAD~\cite{8926407}, and MCP penalties~\cite{8926407} in GTF,
in this paper, we propose the 1st-order GTF model $\cP_0$.

\begin{itemize}
\item We provide theoretical characterizations of the proposed GTF model.
In section 2 we prove that $\cP_0$ is simultaneously performing a k-means clustering on signals $\bm{B}$ and a minimum graph cut on $G$ (Theorem~\ref{thm:0} and Theorem~\ref{thm:1}). 
This means that the model can produce a better result by fusing the information of the node signal with the information of the graph structure to deduce the signal estimate as well as the clustering membership value.

\item In section 3 we propose two algorithms to solve $\cP_0$. 
We first show that the proposed GTF problem is combinatorial in nature, see Theorem~\ref{thm:0} and Theorem~\ref{thm:1}.
We then propose a method to solve the problem efficiently based on the standard technique of spectral approximation.
We finally propose a Hot-bath-based Simulated Annealing algorithm to find the global optimal solution of the proposed GTF problem in case exact solution is required.  

\item In section 4, we apply Theorem~\ref{thm:0} to a semi-supervised learning problem known as the Modified Absorption Problem (MAP)~\cite{pmlr-v38-wang15d,8926407}, to showcase the analysis of the proposed GTF model is generalizable to other models.

\item In section 5 we provide numerical results to support the effectiveness of the algorithms proposed in this work, and to showcase the proposed GTF model outperforms existing GTF models.
In particular, we show that the proposed GTF model has a better performances compared with existing approaches on the tasks of denoising, support recovery and semi-supervised classification. 
We also show that the proposed GTF model can be solved more efficiently than existing models for the dataset with a large edge set.
\end{itemize}

\paragraph*{Notation}
We use $\{$small italic, capital italic, small bold, capital bold$\}$ font to denote $\{$scalar, set, vector, matrix$\}$, respectively.
We focus on simple (without multi-edge) connected unweighted undirected graph $G(V,E)$ of $n$ nodes with signal $\bm{Y}=[\bm{y}_1, \dots, \bm{y}_n]^\top  \in \IR^{n\times d}$ on $G$.
We let $\bm{A} \in \{0,1\}^{n\times n}$ be an adjacency matrix to present the connections between nodes in $G$, with $A_{ij}=1$ indicating nodes $i$ and $j$ are connected and $A_{ij}=0$ otherwise.
We let $\bm{D}$ be a diagonal matrix with the degree of each node on its diagonal.
Let $\bm{L}$ be the Laplacian of $G$, we have $\bm{L}=\bm{D}-\bm{A}$. 
See \cite{Chung:1997} for more background on graph theory.
We let $[n]$ denotes integers from $1$ to $n$.
Let $\bm{X}\in \{0,1\}^{n\times k}$ be the assignment matrix, which can be used to encode the cluster membership of $\{\bm{y}_1, \dots, \bm{y}_n\}$ and the node partition of $G$.

\textit{Remark:}~~
In this work, we focus on undirected graphs.

\section{Equivalent formulation of \texorpdfstring{$\cP_0$}{}}
In this section we provide theoretical characterizations of the proposed GTF model.
We show that the proposed GTF model is equivalent to a simultaneous k-means and min-cut (Theorem~\ref{thm:0} and Theorem~\ref{thm:1}), which means that the model is capable of fusing the information of the node signal with the information of the graph structure to deduce the signal estimate as well as the clustering membership.

\subsection{Theoretical characterizations of the proposed GTF model}
Theorem~\ref{thm:0} shows that the effect of $\mathcal{L}(\bm{B}) > 0$ in $\cP_0$ is to cut the graph $G$ into $k$ partitions and enforce the node signal in these $k$ partitions to be homogeneous (i.e., they share the same signal value).

%
% equivalent formulation
%
\begin{theorem}[The effect of $\mathcal{L}(\bm{B}) > 0$ in $\cP_0$]\label{thm:0}
Given a data matrix $\bm{B} \in \IR^{n \times d}$ over the nodes $V$ of a simple connected unweighted undirected graph graph $G(V,E)$, if  $\mathcal{L}(\bm{B}) > 0$, then it indicates the existence of an edge subset
$$ S_{=1} \coloneqq \Big\{ (i,j) \in E : \mathbb{I}_{\neq 0}\big( \|\bm{b}_i-\bm{b}_j\|_2\big)=1 \Big\}$$ that cuts $G$ into $k$ disjoint partitions $C=\{C_1, \dots, C_k\}$. 
Furthermore, $\mathcal{L}(\bm{B}) > 0$ enforces the node signals in those $k$ partitions sharing the same value. 
Encoding such partition by an assignment matrix $\bm{X}\in \{0,1\}^{n\times k}$, then 
\begin{subequations}
\begin{align}
\mathcal{L}(\bm{B})  
&= \frac{1}{2} \Tr(\bm{X}^\top \bm{L}\bm{X}),
\label{eq:thm1_1}
\\
\bm{B}
&= \bm{X}\bm{\mu},
\label{eq:thm1_2}
\end{align}
\end{subequations}
where $\bm{\mu}=[\bm{\mu}_1, ..., \bm{\mu}_k]^\top \in \IR^{k\times d}$ with $\bm{\mu}_i$ is the node signal for all the nodes in the $i$th partition, and $\bm{L}$ is the graph Laplacian of $G$.
\end{theorem}
{The proof of Theorem~\ref{thm:0} can be found in the Appendix~\ref{append:0}.} 

%
% k-means & graph cut
%
Using Theorem~\ref{thm:0}, we now show that the solution to $\cP_0$ is simultaneously a k-means clustering and minimum graph cut.
\begin{theorem}[$\cP_0$ = k-means and min-cut]\label{thm:1}
 $\cP_0$ is same as
\begin{equation}\label{eq2}
\cP_1 ~:~
\argmin_{ \substack{k\in \mathbb{Z}^+ \\ (\bm{X},\bm{\mu})\in \Psi}}
\frac{1}{2}
\underbrace{\|\bm{Y}-\bm{X}\bm{\mu} \|_F^2}_{\textnormal{k-means} }+\frac{\lambda}{2} \underbrace{\Tr(\bm{X}^\top \bm{LX})}_{\textnormal{graph $k$-cut}}
,
\end{equation} 
where $\Psi \coloneqq \Big\{ \bm{X},\bm{\mu} \Big|
\bm{\mu} \in \IR^{k\times d}, ~ \bm{X}\in \{0,1\}^{n\times k}, ~  \bm{X} \bm{1}_k=\bm{1}_n \Big\}$ and $\bm{1}_k \in \IR^k$ denotes vector of 1s.
\end{theorem}
\begin{proof}
If $\mathcal{L}(\bm{B}) > 0$, apply~\eqref{eq:thm1_1} and \eqref{eq:thm1_2} in  Theorem~2.1 to substitute $\mathcal{L}(\bm{B})$ and $\bm{B}$ in $\cP_0$. 
If $\mathcal{L}(\bm{B}) =0$, it is trivial $\cP_0 = \cP_1$ when $k=1$. 
\end{proof}

\subsection{Simultaneous k-means and graph cut}
The cost function in $\cP_1$ has two terms:
\begin{enumerate}
\item The first term is in the form of k-means clustering on $\bm{Y}$ and $\bm{X}$ is the cluster assignment.
This clustering depends solely on the data $\bm{Y}$ and ignores the graph $G$.
\item The second term is in the form of minimum graph cut on $G$ and $\bm{X}$ is the graph cut assignment matrix.  
This cut depends solely on the graph $G$ and ignores the data $\bm{Y}$.
\item Both k-means and graph cut share the same $\bm{X}$, indicating $\cP_1$ aims to find a $k$-partition that trades off between k-means clustering (using only the signal on $\bm{Y}$) and minimum graph cut (using only the topology on $G$).
This means the information from the signal data and the information of the graph structure are regularizing each other when we solve for $\bm{B}$, in which this forms the basis why model $\cP_0$ is worth studying:
\begin{enumerate}[label=$-$,leftmargin=*,align=left]
\item if the edge structure is consistent with the node data, we are reinforcing the k-means on the node data using the information of the edge;
\item if the edge structure is inconsistent with the signals on nodes, we are correcting (regularizing) the k-means on the node data using the information of the edge.
\end{enumerate}
The above arguments provide explanations why the proposed model $\cP_0$ gave the best recovery in the toy example in Fig.~\ref{fig-toy}.
The model $\cP_0$ utilizes the two attributes of a graph signal: the graph network structure (collective information) and the data on each node (individual information).
The model fuses the two type of information to determine the value of $\bm{B}$.

{\color{black} Note that k-means and graph cut regularizing each other also improves the stability of the model.
For example, in case the graph topology is inaccurate (illustrated by having an inaccurate graph Laplacian $\bm{L}$), then the model is able to use the signal information to correct the inaccuracy of the topology.
Vice versa, if the node signal is highly contaminated (illustrated by the data $\bm{Y}$ is corrupted by noise), then the model is able to make use of the graph topolgy to correct the inaccuracy of the data.
}
\end{enumerate}

We now see that model $\cP_0$ is basically a mixed-integer program, which is NP-hard to solve:
\begin{itemize}
    \item We showed that $\cP_1$ is equivalent to $\cP_0$. 
    \item {\color{black}Theorem~\ref{thm:1} shows that $\cP_1$ is equivalent to solving k-means.
    As k-means is an NP-hard problem~\cite{10.1007/978-3-642-00202-1_24}, thus $\cP_1$ also contains an NP-hard part in its formulation.
    \item Furthermore, Theorem~\ref{thm:1} also shows that $\cP_1$ is
    equivalent to the minimum $k$-cut, which is polynomial time solvable for fixed $k$ \cite{books/daglib/0004338}; however, note that $k$ is also a variable in the formulation of $\cP_1$, and therefore such graph cut problem is also NP-hard \cite{books/daglib/0004338}.
    }
\end{itemize}

\section{The algorithms to solve $\cP_1$}
In the last section we see that problem $\cP_0$ is basically a mixed-integer program $\cP_1$,
where the integer $k \geq 1$ (the number of clusters / partitions) is a unknown in the formulation.
This number can also be treated as the factorization rank in the k-means factorization $\bm{B} = \bm{X} \bm{\mu}$.
It is generally hard to find the optimal $k$ exactly and thus in principle it is challenging to find optimal exact solution for $\cP_1$ (hence $\cP_0$).
In this section, we develop two algorithms to solve $\cP_1$ under two end-user scenarios:
\begin{itemize}
    \item For an approximate solution: a fast approach using spectral approximation.
    \item For accurate solution: a slower approach based on hot-bath simulated annealing.
\end{itemize}

\subsection{Spectral approximation}
To solve problem $\cP_1$ efficiently, we propose a spectral method based on the work of Mark Newman \cite{Newman2006}. 
Here is the idea of the spectral approximation: first we simplify $\cP_1$ to $\cP_2$ by Lemma \ref{cor:1} to eliminate $\bm{\mu}$.
Then we find an approximate solution $(k,\bm{X})$ for $\cP_2$ using brute-force on $k$ and spectral method as follows: 
\begin{itemize}
    \item We first consider $\cP_2$ with a fixed $k$, denoted as $\cP_2^k$.
    \item We solve $\cP_2^k$ to get $(\bm{X}_k, \bm{\mu}_k)$ by the approximate spectral method (Theorem~\ref{thm:vpp} and Algorithm~\ref{algo:spectral}).
    \item We screen through different values of $k$ to find the optimal approximate solution $(k_*,\bm{X}_*)$ for $\cP_2$ (Algorithm~\ref{algo:screenk}).
\end{itemize}
Note that brute-force search on $k$ is not scalable, but we argue that in practice $k$ is usually small, and we can always narrow down the search scope of $k$ using prior knowledge from the application.

\paragraph*{On the diagonal of the matrix $\bm{X}^\top\bm{X}$}
Before we proceed, we recall a useful fact.
For $\bm{X}$ being a 0-1 assignment matrix, the matrix $\bm{X}^\top\bm{X}$ is a diagonal matrix that its $i$th diagonal element is the size of the $i$th cluster, denoted as $|C_i|$.
That is
\begin{equation}\label{eqn:XtX}
\renewcommand{\arraystretch}{1.25}
\bm{X}^\top\bm{X}
~=~
\begin{bmatrix}
|C_1| 
\\
& \ddots 
\\
& & | C_k |
\end{bmatrix}.
\end{equation}
And $\Tr (\bm{X}^\top\bm{X}) =\sum_{i=1}^k |C_i| = n.$

\subsubsection{Eliminate \texorpdfstring{$\bm{\mu}$}{}}
By observation, the variable $\bm{\mu}$ in $\cP_1$ has no constraint.
We simplify $\cP_1$ by eliminating $\bm{\mu}$ using $\bm{\mu}_* = \bm{X}^\dagger \bm{Y}$, where $\bm{X}^\dagger = (\bm{X}^\top \bm{X})^{-1}\bm{X}^\top  $ is the left pseudo-inverse of $\bm{X}$.
This gives the following lemma:
\begin{lemma}[Simplifying $\cP_1$]\label{cor:1}
$\cP_1$ is equivalent to
\begin{equation}\label{prob:pstar}
\cP_2 ~:~
\argmax_{\substack{k\in \mathbb{Z}^+ \\ \bm{X}\in \Phi}} ~ q  ~\coloneqq ~
\Tr \Big(\bm{YY}^\top  \bm{X} \bm{X}^\dagger 
\Big) 
+ \lambda \Tr\Big( \bm{X}^\top  (-\bm{L})\bm{X} \Big)
,
\end{equation} 
where $\Phi=\Big\{
\bm{X} |
\bm{X}\in \{0,1\}^{n\times k}, \bm{X} \bm{1}_k=\bm{1}_n  
\Big\}$ 
and $\bm{X}^\dagger$ denotes pseudoinverse, i.e.,
$\bm{X}^\dagger = (\bm{X}^\top \bm{X})^{-1}\bm{X}^\top $.
\end{lemma}
\begin{proof}
The subproblem on $\bm{\mu}$ in $\cP_1$ has a  closed-form solution $\bm{\mu}^*=(\bm{X}^\top\bm{X})^{-1}\bm{X}^\top \bm{Y}$. 
Then,
\[
\begin{array}{ll}
\hspace{-2mm}\|\bm{Y}-\bm{X}\bm{\mu}^* \|_F^2  
=
\|\bm{Y}-\bm{X}\bm{X}^\dagger \bm{Y} \|_F^2
\\\hspace{-2mm}
=
\| \bm{Y} \|_F^2
-
2 \Tr \Big(
\bm{Y}^\top \bm{X}\bm{X}^\dagger \bm{Y} 
\Big)
+
\Tr \Big(
\big( \bm{X}\bm{X}^\dagger \bm{Y}  \big)^\top \bm{X}\bm{X}^\dagger \bm{Y} 
\Big)
\\

\hspace{-2mm}
=
\| \bm{Y} \|_F^2
-
2 \Tr \Big(
\bm{Y} \bm{Y}^\top \bm{X}\bm{X}^\dagger 
\Big)
+
\Tr \Big(
\bm{Y} \bm{Y}^\top 
( \bm{X}\bm{X}^\dagger )^\top \bm{X}\bm{X}^\dagger 
\Big)
\\

\hspace{-2mm}
=
\| \bm{Y} \|_F^2
-
\Tr \Big(
\bm{Y} \bm{Y}^\top \bm{X}\bm{X}^\dagger 
\Big)
\end{array}
\]
where the last equality is by expanding $( \bm{X}\bm{X}^\dagger )^\top \bm{X}\bm{X}^\dagger $ using $\bm{X}^\dagger = (\bm{X}^\top \bm{X})^{-1}\bm{X}^\top  $ and note that $\bm{X}^\top \bm{X}$ is symmetric (see \eqref{eqn:XtX}).
Now $\cP_1$ is equivalent to
\[
\argmin_{ \substack{k\in \mathbb{Z}^+ \\   \bm{X}\in \{0,1\}^{n\times k} \\ \bm{X}\bm{1}_k = \bm{1}_n  }}
\frac{1}{2}
\| \bm{Y} \|_F^2
-
\frac{1}{2}
\Tr \Big(
\bm{YY}^\top \bm{X}\bm{X}^\dagger 
\Big)
+
\frac{\lambda}{2} \Tr \big(\bm{X}^\top \bm{LX} \big)
,
\]
Multiply the whole expression by 2, ignore the constant, flip the sign, move the negative sign into $\bm{L}$ 
and define $\Phi=\big\{
\bm{X} \,\big|\,
\bm{X}\in \{0,1\}^{n\times k}, \bm{X} \bm{1}_k=\bm{1}_n  
\big\}$ 
gives $\cP_2$.
\end{proof}

\subsubsection{\texorpdfstring{${\cP_2^k}$}{} is approximately a vector partition problem}\label{sec:algo:subsec:approximate}
Let $\cP_2$ with a fixed $k$ be $\cP_2^k$.
By inspecting the cost function $q$ of $\cP_2^k$, we see that $\bm{YY}^\top$ and $\bm{L}$ are positive semi-definite, 
which motivates us to use spectral method to approximate the solution of $\cP_2^k$, based on the following arguments:
\begin{itemize}
    \item in practice, the data matrix $\bm{Y}$ is often low-rank or approximately low rank \cite{udell2019big}; and
    \item in practice, there are only a few clusters.
\end{itemize}
Therefore, we can approximate $q$ by only keeping the terms that are associated with the top $k$ largest eigenvalues of $\bm{YY}^\top$ (and the top $k$ smallest eigenvalues of $\bm{L}$).
To proceed, below we define two sets of spectral vectors.

\paragraph*{Two sets of spectral vectors}
Let the eigendecomposition of $\bm{YY}^\top$ and $\bm{L}$ be 
$\bm{U}\bm{\Sigma}\bm{U}^\top$ and $\bm{V}\bm{\Gamma}\bm{V}^\top$, respectively.
That is, $(\bm{u}_i , \sigma_i)$ denotes the $i$th largest (eigenvector, eigenvalue) of $\bm{YY}^\top$, and similarly for $(\bm{v}_i , \gamma_i)$.
Now, given a constant $\alpha \in \IR$, define two sets of $k$-dimensional vectors $\{\bm{r}_i\}_{i=1}^n$, $\{\bm{t}_i\}_{i=1}^n$ as
\begin{align}\label{eq:r&t}
\hspace{-1mm}
r_i(j) = \sqrt{\sigma_j}U_{ij}, ~~
t_i(j) = \sqrt{\alpha-\gamma_j}V_{ij}, ~~
i \in [n]
\end{align}
where $r_i(j)$ denotes the $j$th element of $\bm{r}_i$.
Note that we will select $\alpha$ such that $\alpha > \gamma_j$ so that $\bm{t}_i$ is a real vector.
We will discuss how to pick $\alpha$ in \cref{alpha}.

Theorem~\ref{thm:vpp} shows that maximizing $q$ in $\cP_2$ is (approximately) equivalent to jointly assigning vectors $\{\bm{r}_i\}_{i=1}^n$ and $\{\bm{t}_i\}_{i=1}^n$ into groups so as to maximizing the magnitude of their weighted sum.
Such an assignment problem is known as a vector partition problem \cite{Newman2006,Alpert1999} and it can be effectively solved by any k-means package.
We remark that the derivation technique of Theorem~\ref{thm:vpp} is based on the work of Mark Newman \cite{Newman2006}.

\begin{theorem}[\texorpdfstring{${\cP_2^k} \approx$}{} Vector partition]
\label{thm:vpp}
Given an integer $1 \leq k \leq n$, a constant $\alpha > \gamma_k$, let $|C_i|$ be the size of the partition $C_i$, then 
$\cP_2^k$ is approximately the following vector partition problem
\begin{equation}
\begin{aligned}
\displaystyle 
\argmax_{\bm{r}_i, \bm{t}_i} ~&
~
\displaystyle 
\sum_{i=1}^k \| \bm{\xi}_i \|^2_2 
+ 
\|\bm{\zeta}_i \|^2_2 
- 
\lambda \alpha n
\\
\textrm{s.t.} ~~~~&
~
\displaystyle  \bm{\xi}_i = \sum_{i\in C_i} \frac{1}{\sqrt{|C_i|}}\bm{r}_i, 
~
\displaystyle \bm{\zeta}_i = \sum_{i\in C_i} \sqrt{\lambda} \bm{t}_i.
\label{prob:vpp}
\end{aligned}
\tag{VPP}
\end{equation}
\end{theorem}

\begin{proof}
The proof has three parts: (1) rewriting $q$ in $\cP_2$ using eigendecomposition, 
(2) define two sets of spectral vectors and (3) approximate the problem as a vector partition problem.
\paragraph*{Rewriting $q$ using eigendecomposition}
Let the eigendecomposition on $\bm{YY}^\top $ and $\bm{L}$  be $\bm{U} \bm{\Sigma} \bm{U}^\top$ and $\bm{V}\bm{\Gamma} \bm{V}^\top$, respectively.
That is, $\bm{U}=[\bm{u}_1, \bm{u}_2, \dots]$ stores the eigenvectors of $\bm{YY}^\top $ and $\bm{\Sigma}$ is a diagonal matrix storing the eigenvalues $\sigma_i$ of $\bm{YY}^\top $,
and likewise for $\bm{V}$, $\bm{\Gamma}$ for $\bm{L}$.
Now in $\cP_2$ we have 
$
q = 
\Tr\big(
    \bm{U} \bm{\Sigma} \bm{U}^\top \bm{X} \bm{X}^\dagger 
    \big)
    + 
    \lambda \Tr\big (
    \bm{X}^\top \bm{V}(-\bm{\Gamma}) \bm{V}^\top \bm{X}
    \big )
$.
Consider $-\bm{\Gamma} = \alpha \bm{I} -\bm{\Gamma} + \alpha \bm{I}$ under a predefined $\alpha$, we have
\[
\begin{array}{rcl}
q  
&=&
\begin{matrix}
\Tr\Big (
\bm{U} \bm{\Sigma} \bm{U}^\top \bm{X} (\bm{X}^\top \bm{X})^{-1}\bm{X}^\top 
    \Big )
    \qquad\qquad\quad
    \vspace{-4mm}
\\
+
\lambda \Tr\Big (\bm{X}^\top \bm{V}(\alpha \bm{I} -\bm{\Gamma}) \bm{V}^\top \bm{X}
    \Big )
- 
\lambda \alpha n
\end{matrix}
\vspace{1mm}
\\
&=&
\begin{matrix}
\Tr\Big (
(\bm{X}^\top \bm{X})^{-1}
\bm{X}^\top 
\bm{U} \bm{\Sigma} \bm{U}^\top \bm{X} 
    \Big ) 
    \qquad\qquad\quad
    \vspace{-4mm}
\\
+
\lambda \Tr\Big (
(\alpha \bm{I} -\bm{\Gamma}) 
\bm{V}^\top \bm{X}
\bm{X}^\top \bm{V}
    \Big )
- 
\lambda \alpha n.
\end{matrix}
\end{array}
\]
Now, invoke \eqref{eqn:XtX},
\[
\renewcommand{\arraystretch}{1}
\begin{array}{rl}
q &= \Tr\left ( 
\begin{bmatrix}
\frac{1}{|C_1| }
\\
& \ddots 
\\
& & \frac{1}{|C_k| }
\end{bmatrix}
\bm{X}^\top 
\bm{U} \bm{\Sigma} \bm{U}^\top \bm{X} 
\right.
\\
&
~\quad + \left. \lambda 
\begin{bmatrix}
\alpha - \gamma_1
\\
& \ddots 
\\
& & \alpha - \gamma_n
\end{bmatrix}
\bm{V}^\top \bm{X}
\bm{X}^\top \bm{V}
 \right )
- 
\lambda \alpha n.
\end{array}
\]

Since trace operator focuses on the diagonal, we only need to look at  the diagonal of both $\bm{X}^\top 
\bm{U} \bm{\Sigma} \bm{U}^\top \bm{X} $ and $\bm{V}^\top \bm{X}
\bm{X}^\top \bm{V}$.
Let $[\bm{M}]_{ii}$ denotes taking only the diagonal of $\bm{M}$ and setting the rest to zero, the trace term of $q$ is equivalent to 
\[
\renewcommand{\arraystretch}{1}
\begin{array}{c}
\Tr\left(
\begin{bmatrix}
\frac{1}{|C_1| }
\\
& \ddots 
\\
& & \frac{1}{|C_k| }
\end{bmatrix}
[\bm{X}^\top \bm{U}]_{ii} [\bm{\Sigma}]_{ii} [\bm{U}^\top \bm{X}]_{ii}
\right.
\\
\qquad\quad
\left.
+  \lambda 
\begin{bmatrix}
\alpha - \gamma_1
\\
& \ddots 
\\
& & \alpha - \gamma_n
\end{bmatrix}
[\bm{V}^\top \bm{X}]_{ii}
[\bm{X}^\top \bm{V}]_{ii}
    \right ),
\end{array}
\]
where we note that this expression is the motivation behind the two sets of spectral vector defined in \eqref{eq:r&t}.

Now let $h = 1,\dots,k$ be the index for cluster $C_h$, 
let $j = 1 ,\dots, n$ be the index of $\sigma_j$ and $\gamma_j$. 
For the matrix product $\bm{C} = \bm{A}^\top\bm{B}$, recall that $C_{jh} = \sum_i A_{ij}B_{ih}$ and note that $\bm{A}^\top\bm{B}$ and $\bm{B}^\top\bm{A}$ share the same diagonal, now $q$ becomes
\[
\begin{array}{rcl}
q
&=&
\displaystyle
\sum_{h=1}^k \frac{1}{|C_h|}
\sum_{j=1}^n\sigma_j \left  (\sum_{i=1}^n U_{ij}X_{ih}\right )^2     
\\
&&
\displaystyle
+ \lambda \sum_{h=1}^k  \sum_{j=1}^n (\alpha-\gamma_j)
\left (\sum_{i=1}^n V_{ij}X_{ih}\right )^2    
    - \lambda \alpha n,
\end{array}
\]
which gives
\begin{equation}
\begin{array}{rl}
q 
&=
\displaystyle
\sum_{h=1}^k \frac{1}{|C_h|} 
\sum_{j=1}^n 
\left  (\sum_{i=1}^n \sqrt{\sigma_j} U_{ij}X_{ih}\right )^2     
\\
&
~~~~\displaystyle
+ \lambda \sum_{h=1}^k  \sum_{j=1}^n 
    \left (\sum_{i=1}^n  \sqrt{\alpha-\gamma_j} V_{ij}X_{ih}\right )^2
- \lambda \alpha n.
\end{array}
\label{proof_eq:Qe1}
\end{equation}

\paragraph*{Define two sets of spectral vectors}
Under a fixed $k$, we now approximately maximize $q$ by focusing on the terms in $q$ that are related to first $k$ largest eigenvalues of $\bm{YY}^\top $ and $-\bm{L}$. 
First we follows \eqref{eq:r&t} and define two sets of $k$-dimensional vectors $\{\bm{r}_i\}_{i=1}^n$ and $\{\bm{t}_i\}_{i=1}^n$.
Note that we pick $\alpha > \gamma_k$ (the $k$th largest eigenvalues of $\bm{L}$) so $\bm{t}_i$ are real for the first $k$ terms.

\paragraph*{Approximate the problem as a vector partition problem}
Now we apply $\{\bm{r}_i\}_{i=1}^n$ and $\{\bm{t}_i\}_{i=1}^n$ on \eqref{proof_eq:Qe1}: in the first term, we drop the terms related to the smallest $n-k$ eigenvalues of $\bm{YY}^\top $; in the second term, we drop the terms related to the smallest $n-k$ of the factors $\alpha - \gamma_i$, then
\begin{align}
q~
&\approx&&
\sum_{h=1}^k \frac{1}{|C_h|} 
\sum_{j=1}^k
\Big  (\sum_{i=1}^n \sqrt{\sigma_j}U_{ij}X_{ih}\Big )^2 
\nonumber
\\
&&&~~~~+
\lambda \sum_{h=1}^k\sum_{j=1}^k\Big (\sum_{i=1}^n \sqrt{\alpha-\gamma_j}V_{ij}X_{ih}\Big )^2 - \lambda \alpha n 
\nonumber
\\
&=&&\sum_{h=1}^k \frac{1}{|C_h|} \sum_{j=1}^k \left (\sum_{i\in C_h} r_{ij}\right )^2
\nonumber
\\
&&&~~~~+ \lambda \sum_{h=1}^k\sum_{j=1}^k\left ( \sum_{i\in C_h} t_{ij}\right )^2 - \lambda \alpha n
\nonumber
\\
&=&& \sum_{h=1}^k 
\underbrace{\sum_{j=1}^k \left (\frac{1}{\sqrt{|C_h|} }\sum_{i\in C_h} r_{ij}\right )^2 }_{\|
\bm{\xi}_h\|_2^2}
\nonumber
\\
&&&~~~~
+ \sum_{h=1}^k
\underbrace{
\sum_{j=1}^k\left ( \sqrt{\lambda}\sum_{i\in C_h} t_{ij}\right )^2}_{ \| \bm{\zeta}_h \|_2^2}
- \lambda \alpha n
\nonumber
\\
&=&&
\sum_{h=1}^k \left \|\bm{\xi}_h\right\|^2 + \left \|\bm{\zeta}_h\right\|^2 - \lambda \alpha n.
\label{proof_eq:Qf}
\end{align}

Changing the index notation $h$ to $i$ in  \eqref{proof_eq:Qf} completes the proof.
\end{proof}

We follow~\cite{Newman2006} to solve \ref{prob:vpp} by applying k-means on the vectors
\[
\bm{z}_i \coloneqq 
\Big [ \dfrac{1}{\sqrt{|C_i|}}\bm{r}_i^\top , \sqrt{\lambda}\bm{t}_i^\top  \Big ]^\top
\]
for $i \in [n]$.
This gives an approximate solution of $\cP_2^k$. 
See Algorithm~\ref{algo:spectral}. 

{\color{black}
\textit{When does the spectral approximation work?}
The above derivation shows that the estimation of $k$ plays an important role in the whole process.
We note that such an approach is effective when the true $k$ is small, which is grounded on the fact that most of the information is concentrated in a few eigenvalues of the data covariance matrix $\bm{Y}\bm{Y}^\top$ and a few eigenvalues of the Laplacian $\bm{L}$. 
A small true $k$ also makes the algorithm effective: the spectral approximation relies on the approximation of a sum with $n$ terms in \eqref{proof_eq:Qe1} by the sum with $k$ terms in \eqref{proof_eq:Qf}, where $n$ is the number of nodes that is potentially  large.
In the case $k \ll n$, the spectral approximation dramatically reduces the computational complexity of the whole algorithm and making the brute force search on $k$ possible.
See Fig.~\ref{fig-sim1} in the experiment for an example where $k=4 \ll 2642=n$.

If it is the case that the true $k$ is large, spectral approximation no longer works.
However, we argue that $k \ll n$ is true most of the time in practice, see discussion in Section~\ref{sec:algo:subsec:approximate}.
}

\begin{algorithm}
\caption{Spectral method for solving $\cP_2^k$}
\begin{algorithmic}
\STATE \textbf{Input}: $\bm{Y}\in\IR^{n\times d}$, $\bm{L}\in \IR^{n\times n}$, $\lambda\in \mathbb{R}$, and $k$.
\vspace{1mm}

\STATE \textbf{Compute}: 
\bindent
\vspace{1mm}

\STATE \hspace{-5mm} Top $k$ largest eigenvalues
$\{ \sigma_i \}_{i=1}^k$ 
and the corresponding eigenvectors $\{ \bm{u}_i \}_{i=1}^k$ of $\bm{YY}^\top $;
\vspace{1mm}

\STATE\hspace{-5mm}  Top $k$ smallest eigenvalues  $\{ \gamma_i \}_{i=1}^k$ and the corresponding eigenvectors $\{ \bm{v}_i \}_{i=1}^k$ of $\bm{L}$;
\vspace{1mm}

\STATE \hspace{-5mm} $\displaystyle \alpha = \frac{1}{n-k}\sum_{i=k+1}^n \hspace{-1mm}\gamma_i$\;
\vspace{1mm}

\STATE \hspace{-5mm} 
$\bm{z}_i \coloneqq 
\Big [ \dfrac{1}{\sqrt{|C_i|}}\bm{r}_i^\top , \sqrt{\lambda}\bm{t}_i^\top  \Big]^\top$ 
by \eqref{eq:r&t}, $i \in [n]$;
\vspace{1mm}

\STATE Run k-means on $\{\bm{z}_i\}_{i=1}^n$ and obtain $\bm{X}.$
\eindent
\vspace{1mm}

\STATE \textbf{Return} $\bm{X}_k=\bm{X}$.
\end{algorithmic}
\label{algo:spectral}
\end{algorithm}

\subsubsection{Approximation error and the choice of $\alpha$}\label{alpha}
In the approximation we drop the $n-k$ most negative eigenvalues of $-\bm{L}$ to approximate the matrix $\alpha \bm{I} -\bm{L}$ (i.e., the $n-k$ leading eigenvalues of $\bm{L}$), and, instead of computing $\bm{V}(\alpha \bm{I} - \bm{\Gamma})\bm{V}^\top $, we compute $\bm{V}(\alpha \bm{I}'-\bm{\Gamma}')\bm{V}^\top $, where $\bm{I}'$ and $\bm{\Gamma}'$ are matrix $\bm{I}$ and $\bm{\Gamma}$ with the last $n-k$ diagonal elements set to zero.
The approximation error in such a procedure is
\[
\begin{array}{rcl}
\varepsilon 
&\coloneqq&
\Tr\Big (\bm{V}(\alpha \bm{I} - \bm{\Gamma})\bm{V}^\top  - \bm{V}(\alpha \bm{I}'-\bm{\Gamma}')\bm{V}^\top  \Big)
\\
&=&
\Tr\Big( (\alpha \bm{I} - \bm{\Gamma}) - (\alpha \bm{I}'-\bm{\Gamma}') \Big) 
\,=\, \displaystyle  \sum_{i=k+1}^n (\gamma_i-\alpha)^2.
\end{array}
\]
Let $\dfrac{\partial \varepsilon}{\partial\alpha}=0$ gives the optimal $\alpha^* = \dfrac{1}{n-k}\displaystyle \sum_{i=k+1}^n\gamma_i$.
\medskip

\subsubsection{The whole approximate spectral method}
We now propose Algorithm~\ref{algo:screenk} to solve $\cP_2$ approximately.

\begin{algorithm}
\caption{Algorithm for solving $\cP_2$ approximately}
\begin{algorithmic}
\STATE \textbf{Input}: $\bm{Y}\in\IR^{n\times d}$, $\bm{L}\in \IR^{n\times n}$, $\lambda\in \mathbb{R}$, and $k_{\max}\in \IZ^+$.
\vspace{1mm}

\STATE \textbf{Initialize}: $q^*=+\infty$.
\vspace{1mm}

\FORALL{$k=1, 2, ... , k_{\max}$}
    \STATE Using Algorithm~\ref{algo:spectral} to get $\bm{X}_k$;
    \vspace{1mm}
    
    \STATE Compute $q_k= \Tr (\bm{YY}^\top  \bm{X}_k \bm{X}_k^\dagger ) + \lambda \Tr\Big( \bm{X}_k^\top (-\bm{L})\bm{X}_k \Big)$;
    \vspace{1mm}

    \STATE Update $q^*=q_k$ and $\bm{X}_*=\bm{X}_k$ if $q_k\leq q^*$.
    \vspace{1mm}
    
  \ENDFOR
\STATE \textbf{Return}  $\bm{X}_*$ and $\bm{\mu}= ( \bm{X}_*^\top \bm{X}_*)^{-1} \bm{X}_*^\top \bm{Y}$.
\end{algorithmic}
\label{algo:screenk}
\end{algorithm}

\subsection{Heat-bath simulated annealing}
{\color{black} 

Theorem~\ref{thm:1} indicates that $\cP_1$ is a NP-complete problem. 
In the previoius section, we propose Algorithm~\ref{algo:screenk} based on approximating the eigen-spectrums of $\bm{Y}\bm{Y}^\top$ and $\bm{L}$. 
In this section, we propose another approach that uses Heat-bath simulated annealing to find the global optimal solution of ${\cP_1}$.

Simulated annealing is a probabilistic optimization technique that is used to find global solutions.
In simulated annealing, we need a Hamiltonian energy function that represents the objective function we aim to minimize.
This energy function maps each solution to a real number representing its ``energy''.
The goal of the algorithm is to find the state with the lowest possible energy. 
We rewrite ${\cP_1}$ into ${\cP_3 }$ as follows and define $\mathcal{H}$ as the Hamiltonian energy function. }
\begin{equation}\label{sa_form}
\begin{aligned}
\cP_3 ~:~ \argmin_{ k, \bm{\delta} } & 
~~\mathcal{H}( \bm{\delta} ) 
\coloneqq
\underbrace{\sum_{i=1}^n\sum_{j=1}^{k}\mathbb{I}_{=j}(\delta_i)\left \|\bm{y}_i-\bm{c}_j\right \|_2^2}_{\textnormal{k-means}}
\\
& \qquad\qquad  + \lambda \underbrace{\sum_{i,j}\sum_{s\ne t}A_{ij}\mathbb{I}_{=s}(\delta_i)\mathbb{I}_{=t}(\delta_j)}_{\textnormal{graph $k$-cut}}
\\
s.t. & 
~~\bm{c}_j = \displaystyle 
\dfrac{\displaystyle \sum_{l=1}^n \mathbb{I}_{=j}(\delta_l)\bm{y}_l }
{\displaystyle\sum_{l=1}^n \mathbb{I}_{=j}(\delta_l)},
\ j \in \{1, 2, ..., k\} ;\\
       & ~~\delta_i \in \{1, 2, ..., k\}, \ i = 1, 2, ..., n,
\end{aligned}
\end{equation} 
where the vector $\bm{\delta} = [\delta_1, \delta_2, ... , \delta_n ]^\top \in \{1,\dots,k\}^n $ is a membership state, with the $i$th element $\delta_i \in [k] \coloneqq \{1,2,,\dots,k\}$ representing the membership of signal $\bm{y}_i$ and the membership of node $i$, 
e.g., $\delta_i = h$ means the  $\bm{y}_i$ and node $i$ belong to cluster $h$.
The symbols $s,t\in [k]$ are two dummy cluster labels and the indicator function $\mathbb{I}_{=j}(\delta_i) = 1$ if $\delta_i=j$, and $= 0$ otherwise.
The matrix $\bm{A}$ with elements $A_{ij} \in \{0,1\}$ is the adjacency matrix of the graph $G$, and lastly, $\bm{c}_j$ is an intermediate variable in $\cP_3$.

It is trivial to see the equivalence between ${\cP_1}$ and $\cP_3$:
in $\cP_3$ we uses $\bm{\delta}$ to replace the assignment matrix $\bm{X}$ in $\cP_1$.
The membership states $\{\delta_i\}$ act as a 0-1 on/off switches in both the k-means term and the graph cut term.

\subsubsection{Contrasting $\cP_1$ and $\cP_3$}
Recall that in Algorithm~\ref{algo:screenk} for (approximately) solving ${\cP_1}$, we screen through different values of $k$ in a pre-defined search set $[k_{\max}]$ to find the optimal $k_*$ and we assume $k_* \in [k_{\max}]$.
For $\cP_3$, we propose to use heat-bath Simulated Annealing (SA) to find $k_*$.
We start with an overestimated value $k$ (which is expected to be larger than $k_*$), then SA will find $k_*$ (probabilistically) by a ``cooling process."

\subsubsection{Heat-bath simulated annealing}
To minimize $\mathcal{H}(\bm{\delta})$, we use a heat-bath Monte Carlo optimization strategy (abbreviated as heat-bath), which is a strategy similar to the classical SA (abbreviated as cSA) but is shown to have a better performance~\cite{Bornholdt2006,Carlon2013}.
While cSA and heat-bath are conceptually similar, cSA is based on the Metropolis approach~\cite{10.2307/2334940} to compute the transition probabilities without computing partition functions because the partition functions are hard to compute for many applications.
However, in the setting of GTF presented in this paper, efficient computation of the partition function is not a challenge, which allows the use of the alternative heat-bath transition probabilities.

\paragraph*{Coordinate-wise update}
In the proposed SA approach to solve $\cP_3$, we perform a coordinate update of $\bm{\delta}$: 
we update one element $\delta_i$ at a time while keeping the rest of $\bm{\delta}$ fixed at their most recent value.

\paragraph*{Details of the heat-bath update}
As $\delta_i \in [k]$ represents a cluster label, the update on $\delta_i$ is a switch process.
Let $s,t\in [k]$ be cluster labels, where $s$ is the state of $\delta_i$ before the update and $t$ is the state of $\delta_i$ after the update.
Such switching update of $\delta_i$ is carried out using single-spin heat-bath update rule~\cite{10.1093/oso/9780198803195.003.0009}:
we update the system energy $\mathcal{H}$ when making a membership assignment $\delta_i$ switch from $s$ to $t$,
i.e.,
\[
\underbrace{\mathcal{H}(\delta_i=t)}_{\text{energy after update}}
=
\underbrace{
\mathcal{H}(\delta_i=s)}_{\text{energy before update}}
+
\underbrace{\Delta \mathcal{H}(\delta_i= s\rightarrow t)}_{\text{change in energy}}
.
\]
The probability of the membership switch at a temperature $T$ is proportional to the exponential of the corresponding energy change of the entire system, i.e
\begin{equation}\label{update}
\begin{aligned}
\mathbb{P}(\delta_i=t;T) 
= 
\dfrac{ 
\exp\left \{ -\dfrac{1}{T} \Delta \mathcal{H}(\delta_i= s\rightarrow t) \right \} 
}
{
\displaystyle 
\sum_{l=1}^{k} \exp\left \{ -\frac{1}{T} \Delta \mathcal{H}(\delta_i= s\rightarrow l) \right \} }.
\end{aligned}
\end{equation}

\paragraph*{A brief summary of how SA finds global minimum}
From statistical mechanics, at a high enough temperature, the membership state $\delta_i$ follows a discrete uniform distribution, 
i.e.,
\[
\begin{array}{l}
\displaystyle
\lim_{T \rightarrow + \infty}
\underbrace{ 
\dfrac{ 
\exp\left \{ -\frac{1}{T} \Delta \mathcal{H}(\delta_i= s\rightarrow t) \right \} 
}
{
\displaystyle 
\sum_{l=1}^{k} \exp\left \{ -\frac{1}{T} \Delta \mathcal{H}(\delta_i= s\rightarrow l) \right \} }
}
_{
\mathbb{P}(\delta_i=t;T) 
}
\vspace{3mm}
\\
~~~=
\dfrac{ 
\displaystyle
\lim_{T \rightarrow + \infty}
\exp\left \{ -\frac{1}{T} \Delta \mathcal{H}(\delta_i= s\rightarrow t) \right \} 
}
{
\displaystyle 
\sum_{l=1}^{k} \lim_{T \rightarrow + \infty} \exp\left \{ -\frac{1}{T} \Delta \mathcal{H}(\delta_i= s\rightarrow l) \right \} 
}
~=~ \cfrac{1}{k} \, ,
\end{array}
\]
hence at such moment, it is capable of jumping out of a local minimum, or in other words, SA at high temperatures is able to explore all the $k$ possible membership assignments.
When the temperature cools down (say $T \rightarrow 1$), the state $\delta_i$ obeys a categorical distribution where $\delta_i$ are more likely to switch to $t$ from the current membership state $s$ if the energy change $\Delta \mathcal{H}(\delta_i= s\rightarrow t)$ is negative and small. 

\paragraph*{The whole SA algorithm}
Based on the above discussion, we propose the SA Algorithm~\ref{alg:sa}, where scalars $T_{start}$ and $T_{end}$ are the start and end temperatures, $c \in [0,1 )$ is a constant to control the cooling rate, 
and $K$ is the number of sweep times for each temperature. 

\begin{algorithm}
\caption{Hot-bath SA for solving $\cP_3$.}
\begin{algorithmic}
\STATE \textbf{Input}: $T_{start}=100$, $T_{end}=0.001$, $c=0.99$, $K$, $\lambda$.
\STATE $T=T_{start}$. 
\vspace{1mm}
\\
    \WHILE{$T\geq T_{end}$}\vspace{1mm}
        \FORALL{$k= 1, 2, ..., K$ }\vspace{1mm}
        \STATE $R = $ RandomShuffle($\{1,2,\dots,n\}$); \vspace{1mm}
        \FORALL{$i=1, 2, ..., n$}\vspace{1mm}
         \STATE Update $\mathcal{H}(\delta_{R[i]})$ at $T$ based on the probability~\eqref{update} using roulette wheel algorithm~\cite{10.1093/oso/9780195099713.001.0001};
         \vspace{1mm}
         \\
         \STATE   Update $\mathcal{H}$ after $\delta_{R[i]}$ is assigned;\vspace{1mm}
        \ENDFOR\vspace{1mm}
        \ENDFOR\vspace{1mm}
    \STATE $T=T \times c$ ~~ \% cool down the temperature.\vspace{1mm}
    \ENDWHILE\vspace{1mm}
    \\
\textbf{Return} The most recent $\bm{\delta}$.
\end{algorithmic}
\label{alg:sa}
\end{algorithm}

The major computational cost here comes from computing the term $\Delta \mathcal{H}(\delta_i= s\rightarrow t)$, which can be efficiently computed in $\mathcal{O}(n)$~\cite{SELIM19911003,Johnson1989}. 
However, we remark that it can take many iterations for the Hot-bath SA algorithm to converge (coold down).

\begin{figure*}[t!]
\begin{center}
\includegraphics[width=\textwidth]{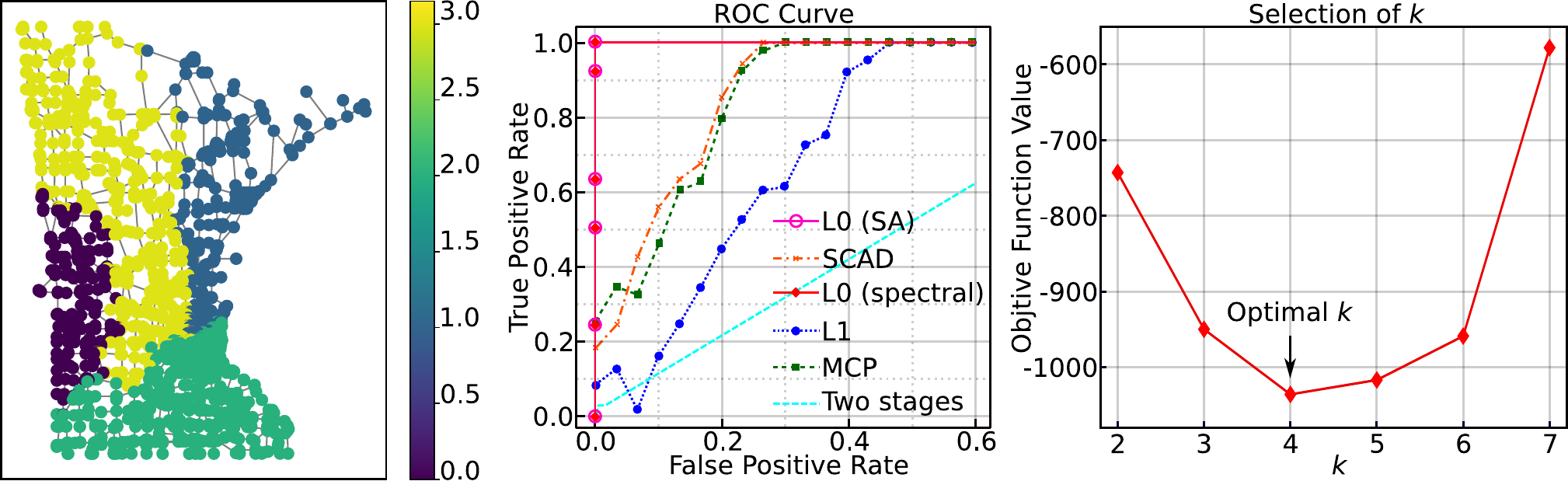}
\caption{On identifying boundary edges. 
Left: the ground truth $\bm{y}_*$ on the Minnesota road graph. 
Middle: the ROC curve of the methods. 
The curves for the spectral method and SA method are overlapped.
Here the $\ell_0$ model achieved perfect ROC.
Right: the optimal $k_*$ ($=4$ clusters) for the $\ell_0$ model $\cP_2$.
}
\label{fig-sim1}
\end{center}
\end{figure*}

\subsection{Section summary}
We recall that the main contribution of this work is the new GTF model $\cP_0$.
In this section we proposed two algorithmic approaches for solving $\cP_0$.
In general, compared with the spectral method, the Hot-bath SA algorithm provides a more accurate solution but it also requires more computational time to find the solution.
We refer to the discussion in \cref{sec_exp} for more details.

\section{Extension to graph-based Transductive Learning}\label{sec_map}

In this section, we show that the Theorem~\ref{thm:0} is generalizable to other models.
Let $K > 0$ be a given integer.
Consider a $K$-class classification problem in a semi-supervised learning setting: given a dataset with $n$ samples, we observe a subset of class labels encoded in a class assignment matrix $\bm{Y}\in\{0,1\}^{n\times K}$, and
a diagonal indicator matrix $\bm{M}\in\{0,1\}^{n\times n}$ that denotes the samples whose labels have been observed.
Such classification problem, called the Modified Absorption Problem~\cite{Talukdar2009,talukdar-pereira-2010-experiments} can be expressed as a 1st-order GTF with $\ell_{2,0}$-norm penalty as
\begin{align}\label{eq:ssl}
\hspace{-2mm}
\mathcal{Q} : \argmin_{\bm{B}\in \IR^{n\times K}} 
\frac{\left \| \bm{M} (\bm{Y}-\bm{B})\right  \|_F^2 }{2}
+ \lambda \cL(\bm{B})
+ \epsilon \| \bm{R}-\bm{B} \|_F^2,
\end{align}
where $\bm{R}\in\IR^{n\times K}$ encodes a fixed prior belief on the samples, and $\lambda \geq 0, \epsilon \geq 0 $ are regularization parameters.
The matrix $\bm{B}\in \IR^{n\times K}$ contains the fitted probabilities, with $B_{ij}$ representing the probability that sample $i$ is of class $j$. 
Using $\bm{b}_i$ to denote the $i$th row of $\bm{B}$, the middle term of~(\ref{eq:ssl}) encourages the probabilities of samples to behave smoothly over a  graph $G(V,E)$ constructed based on the sample features.

Now we give a theorem related to $\mathcal{Q}$.
The focus here is to demonstrate the generality of Theorem~\ref{thm:0}, which can be applied to models with the 1st-order GTF with $\ell_{2,0}$-norm penalty.

\begin{theorem}\label{thm:ssl}
$\mathcal{Q}$ is equivalent to
\begin{equation}
\mathcal{Q}_1 :
{
\argmin_{ 
\substack{k\in \IZ^+
\\
\bm{X} \in \{0,1\}^{n\times k}
\\
\bm{X}^\top \bm{1}_k=\bm{1}_n 
}
}
}
\begin{array}{l}
\dfrac{1}{2} \Big\|
\bm{M}\Big(\bm{Y}-\bm{X}\bm{B}(\bm{X})\Big)
\Big\|_F^2
\vspace{-3mm}
\\ 
~ + 
\lambda \Tr (
\bm{X}^\top \bm{LX}  ) 
+ 
 \epsilon \big\|
\bm{R}-\bm{X}\bm{B}(\bm{X})
\big\|_F^2,
\end{array}
\label{eq:ssl2}
\end{equation}
with $\bm{B}\big(
\bm{X})=(\bm{X}^\top (\bm{M}+\bm{I})\bm{X}\big)^{-1}
\big(
\bm{X}^\top \bm{MY}+\epsilon \bm{X}^\top \bm{R}
\big)$.
\end{theorem}
We use the techniques for solving $\cP_2$ to solve $\mathcal{Q}_1$ (hence solving $\mathcal{Q}$).
Note that $\mathcal{Q}$ and $\cP$ have a different optimization subproblem, so we cannot use Algorithm~\ref{algo:spectral} directly to solve the inner minimization of $\mathcal{Q}$.
We use \textcolor{black}{the Frank-Wolfe algorithm~\cite{https://doi.org/10.1002/nav.3800030109,10.5555/3042817.3042867}} to solve such inner minimization of $\mathcal{Q}$.

\begin{figure*}[!t]
\begin{center}
\includegraphics[width=0.85\textwidth]{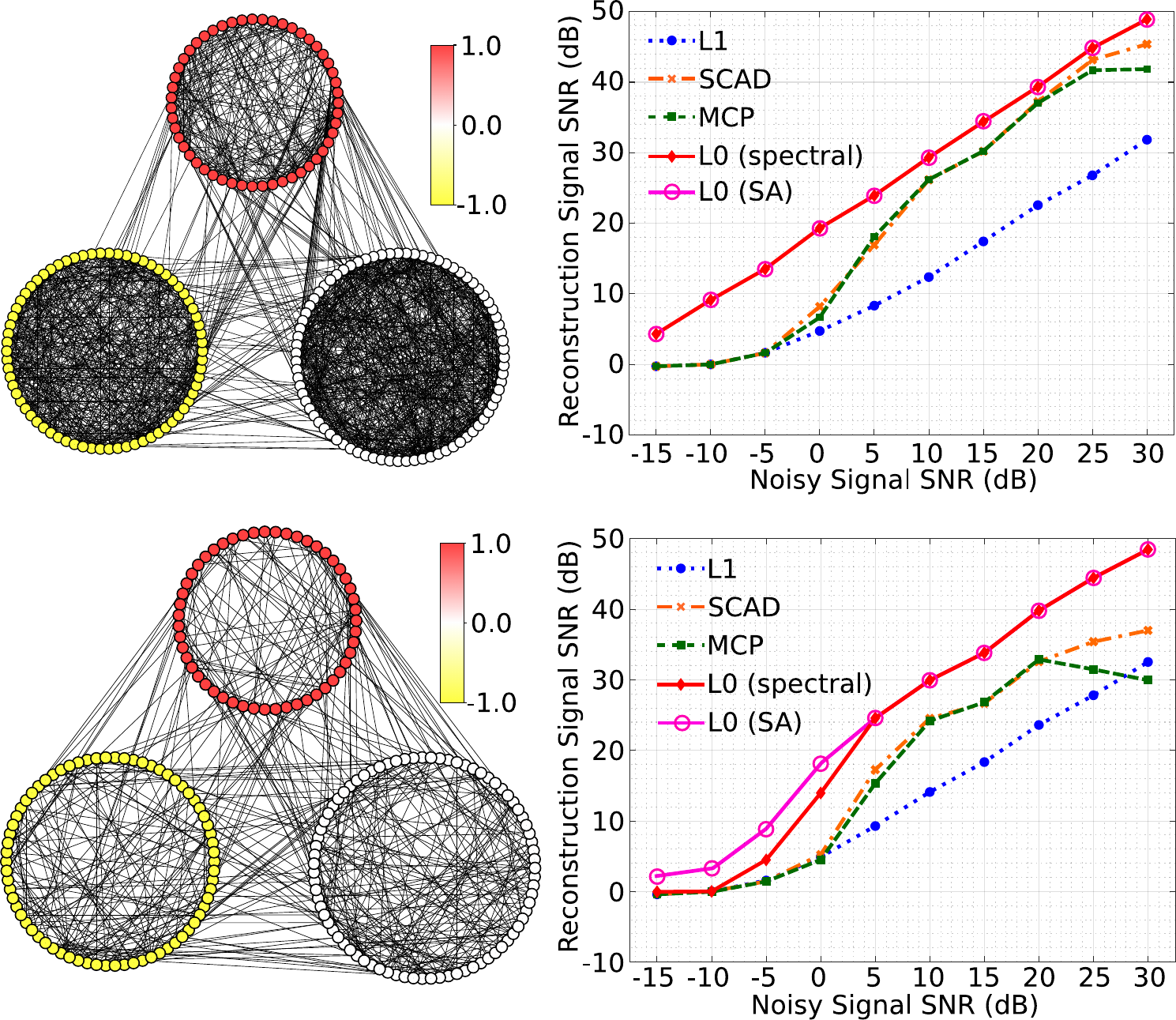}
\caption{
Left: the graph structure of $G_1$ (top) and $G_2$ (bottom) and the corresponding ground truth signal input to each node in both graphs.
Right: the corresponding plots of input signal SNR versus reconstructed signal SNR for the GTF models.
Note that in both graphs, the $\ell_{2,0}$ model consistently outperforms other GFT models.
}
\label{fig-sim2}
\end{center}
%\vspace*{-10mm}
\end{figure*}

\section{Experimental results}\label{sec_exp}
We now demonstrate the proposed 
$\ell_{2,0}$-norm penalized 1st-order GTF model 
(i.e., problem \eqref{prob:1}) outperforms three state-of-the-art GTF models (see table\,\ref{table:gtfmodels}) over two graph learning tasks: support recovery and signal estimation.
Then we show the proposed MAP-GTF model outperforms other models in the task of semi-supervised classification. 
All experiments were run on a computer with 8 cores 3.7GHz Intel CPU, and 32 GB RAM\footnote{The code is available at: \url{https://github.com/EJIUB/GTF_L04}}.

\begin{table}[h]
\setlength\belowcaptionskip{1pt}
\caption{GTF models with different penalty} 
\label{table:gtfmodels}
\centering
\normalsize
\renewcommand{\arraystretch}{1.33}
\begin{tabular}{ l | l}
\hline 
Penalty &  Param. \\
\hline \hline
$\ell_{2,0}$-norm  (this work) & $k,\lambda$\\
$\ell_1$-norm \cite{pmlr-v38-wang15d} &    $\lambda$\\
SCAD \cite{8926407} & $\gamma, \tau, \lambda$\\
MCP  \cite{8926407} & $\gamma, \tau, \lambda$
\end{tabular}
\end{table}
\raggedbottom

\subsection{Support recovery}
\subsubsection{Dataset}
We test all the methods for identifying the boundaries of a piece-wise constant signal on the Minnesota road graph~\cite{nr} shown in Fig.~\ref{fig-sim1}.
The graph $G$ has $n=2,642$ nodes and $3,304$ edges.
Let $\bm{y}_*\in \IRn$ be the ground truth scalar-valued signal for each node in $G$, we add Gaussian noise $\bm{\theta} \sim \mathcal{N}(0,\sigma^2I)$ with $\sigma^2=0.05$ to $\bm{y}_*$ to obtain the simulated observed signal $\bm{y} = \bm{y}_* + \bm{\theta}$.
The Signal-to-Noise Ratio (SNR) of $\bm{y}_*$ is $10\log_{10} \frac{\|\bm{y}_*\|}{\sigma^{2}nd} \approx12$. 

\subsubsection{Support recovery: perfect ROC curve for the proposed method}
The task here is to identify the boundary edges by recovering the true supports of $\left \|\bm{b}_i-\bm{b}_j\right \|_0, \ \forall (i,j) \in E$.
We evaluate the performance of the aforementioned methods by comparing their result with the ground truth boundary edges computed from $\bm{y}_*$ as shown in the left panel of Fig.~\ref{fig-sim1}.

\subsubsection{Parameter tuning}
In the test, we report the best results for each model by tuning their hyper-parameters.
\begin{itemize}
\item We run Algorithm~\ref{algo:screenk} to use the spectral method to solve the proposed $\ell_{2,0}$ model \eqref{prob:1} for the support recovery problem.
The algorithm has two hyper-parameters: the number of partitions $k$ and the regularization parameter $\lambda$. 

\item We run Algorithm~\ref{alg:sa} to use the SA method to solve the proposed $\ell_{2,0}$ model \eqref{sa_form}. 
We set the number of partitions $k = 7$. 
The algorithm has only one hyper-parameter: the regularization parameter $\lambda$. 

\item For the $\ell_1$, SCAD, and MCP models, we use the source code provided by~\cite{8926407} to identify supports for this experiment.
For SCAD, we set $\gamma=3.7$ as suggested in~\cite{8926407}.
For MCP, we set $\gamma=1.4$ as suggested in~\cite{8926407}. 
\end{itemize}

\subsubsection{The results}
We plot the results in Fig.~\ref{fig-sim1}.
We compare the methods in terms of ROC curve, i.e. the true positive rate versus the false positive rate of identification of a boundary edge correctly.
The $\ell_{2,0}$ model solved by both the proposed spectral method and the SA method obtains a perfect ROC curve, which significantly outperforms other models.
The GTF models with nonconvex penalties (SCAD and MCP) have similar performance, which is better than the $\ell_1$ model.

It is not surprising that the proposed $\ell_{2,0}$ model achieves a significantly better result; see the discussion in Remark 2.3.
% by Theorem~\ref{thm:1}, the $\ell_{2,0}$ model is equivalent to a k-means clustering with graph cut regularization, in which running k-means clustering here significantly improves the solution, i.e., achieving a perfect ROC curve. 
% The rationale is that k-means improves the model's ability to adapt to the inhomogeneity of the signal.

In Fig.~\ref{fig-sim1} we also show that we can find the $k_*$ for $\cP_2$ by screening different values of $k$ in \textbf{Algorithm}~\ref{algo:screenk}.
As the search space of $k$ is small, hence the proposed method runs very fast. 
In addition, the SA method can also find the $k_*$ as long as we set  $k$ larger than $k_*$. 

{\color{black}
In Fig.~\ref{fig-sim1}, we also showed that a ``two-stage approach'' for  comparison, where the ``two-stage approach'' refers to first partition the graph and then estimating signals.
As shown by the ROC curves, the two-stage approach is not competitive. 
The reason is that the communities in the graph are inconsistent with the clusters of the signals on the graph.
That is, the communities identified by the community detection algorithm (normalized graph cut algorithm) based on the graph topology do not agree with the clusters identified by the k-means algorithm based on the signal on the nodes. 
Clearly, in this graph the clusters of the signal on nodes provide more insights about the location of the boundaries.
Identifying communities first in the two-stage approach would miss the cluster structures expressed in the signal on nodes and, therefore, mislead the identification of the boundaries. 
This in fact showcases the importance of the proposed model, which is able to perform simultaneous graph clustering and k-means. 
In comparison to the two-stage approach, the proposed approach in this paper balances the information from the community detection on the graph topology and the clustering information on the signal (as shown in Theorem~\ref{thm:1}). Therefore, the proposed approach is better at detecting the boundaries. 
}

\begin{table*}[]
\setlength\belowcaptionskip{1pt}
\caption{Time comparison (mean(std)) with the number of edges in graphs increasing.}\label{tab:time}
\centering
\normalsize
\renewcommand{\arraystretch}{1.75}
\begin{tabular}{ccccccc}
 \hline
         & \multicolumn{2}{l}{($p=0.1, q=0.01$)} & \multicolumn{2}{l}{($p=0.5, q=0.1$)} & \multicolumn{2}{l}{($p=0.9, q=0.2$)} \\ \cline{2-7} 
         & $|E|$           & Time(s)             & $|E|$          & Time(s)             & $|E|$          & Time(s)             \\ \hline
$\ell_1$ & 805(23)         & 4.71(1.22)          & 4707(69)       & 56.74(20.43)        & 8750(40)       & 134.92(46.53)       \\
SCAD     & 805(23)         & 5.22(1.24)          & 4707(69)       & 50.32(11.72)        & 8750(40)       & 106.03(25.34)       \\
MCP      & 805(23)         & 7.71(2.26)          & 4707(69)       & 42.71(14.01)        & 8750(40)       & 105.64(33.92)       \\
Alg.~\ref{algo:screenk} & 805(23)         & 0.01(0.01)          & 4707(69)       & 0.01(0.01)          & 8750(40)       & 0.01(0.01)          \\
Alg.~\ref{alg:sa} & 805(23)         & 990.54(5.42)        & 4707(69)       & 993.14(6.41)      & 8750(40)       & 992.94(8.91)     
\end{tabular}
\end{table*}

\subsection{Denoising using GTF}
We now compare the performance of the GTF models on denoising.

\subsubsection{Dataset: graph generation}\label{graph_gen}
{\color{black} We generate a simple connected undirected unweighted graph $G(V,E)$ with $200$ nodes based on a random graph model called Stochastic Block Model~\cite{HOLLAND1983109}.
} 
We plant three modules/communities in $G$, which have 50 nodes, 70 nodes, and 80 nodes, respectively. 
We introduce two parameters $(p,q)$: 
$p$ controls the probability of connections between nodes within the modules/communities and
$q$ controls the probability of connections between nodes across the modules/communities. 
Then we generate two graphs $G_1, G_2$ as in Fig.\ref{fig-sim2}.
For $G_1$, we set $p=0.2$ and $q=0.05$ and for $G_2$, we set $p=0.05$ and $q=0.01$.
Clearly, the modular/community structure of $G_2$ is relatively less clear than that of $G_1$.

\subsubsection{Dataset: signal generation}\label{signal_gen}
We generate a ground truth signal $\bm{y}_*\in\IR^{200}$ on the graphs $G_1$ and $G_2$ where on the module/community with $50$ nodes has an input signal value of $1$, on the module/community with $70$ nodes has an input signal value of $-1$, and on the module/community with $80$ nodes has input signal value of $0$.
We simulate $d=10$ independent noisy realizations of the graph signal and concatenate them to construct a noisy vector-valued graph signal $\bm{Y}=[\bm{y}_* + \bm{\theta}_1, \bm{y}_* + \bm{\theta}_2, ...]\in \IR^{200\times 10}$, where all $\bm{\theta}_i\in\IR^{200}$ are independently drew from $\theta_i\sim \mathcal{N}(0,\sigma^2I)$.

\subsubsection{The learning task}
The goal here is to use different GTF models to recover the vector-valued signal under different input signal SNR controlled by $\sigma$.
We recall that the reconstructed signal SNR is computed as $10\log_{10}(\|\bm{Y}^*\|_F/\|\bm{B}-\bm{Y}^*\|_F)$, where $\bm{B}$ is the reconstructed signal.

\subsubsection{Parameter tuning}\label{para_set}
We use the same setup of hyper-parameters tuning as in the experiment on support recovery, and for each GTF model, we screen through the hyper-parameters and report the best result.

\subsubsection{The results}
In Fig.\ref{fig-sim2}, we plot the SNR of the reconstructed signal on the graphs $G_1$ (top) and $G_2$ (bottom)
across different SNR values of the input signal for all the GTF models. 
The curve of the $\ell_{2,0}$ model solved by the proposed method is above all the other GTF models, which demonstrates that our model can take advantage of the graph structure to recover the true signal.
For the reconstruction of signal on the graph $G_2$, which is a harder problem as the modular/community structure of $G_2$ is less clear than $G_1$,  the $\ell_{2,0}$ model still has a better reconstruction SNR than all other GTF models, with a consistent amount of at least 5dB better SNR in the high input SNR regime. Furthermore, we find that the SA method has better performance than the spectral method when the noisy signal SNR is smaller for signal recovery on $G_2$.

\subsection{Computational time}
In this section, we compare the computational time of the competing GTF models. 

\subsubsection{Dataset} 
We follow the procedures described in sections~\ref{graph_gen} and~\ref{signal_gen} to generate graph signals for time comparison. 
The number of nodes and the community structure of the graphs are the same as described in section~\ref{graph_gen}.
All generated graphs have 200 nodes and 3 communities with 50, 70, and 80 nodes, respectively. 
We control the number of edges of the graphs by setting different $(p, q)$ pairs (specified in Table~\ref{tab:time}). 
We follow the procedure in section~\ref{signal_gen} to generate the signal on graphs. 

\subsubsection{Parameter}
For each GTF model, we use the best parameters obtained in section~\ref{para_set}. 

\subsubsection{The results}
In Tabel~\ref{tab:time}, we show the time comparison for different GTF models on denoising graph signals. 
For each  $(p, q)$ pair, we generate 10 graph signals with 10 different graphs, which have the same community structure but different numbers of edges. 
We show the average and standard deviation of computational time for each GTF model. 
We observe that the computational times of previous GTF models ($\ell_1$, SCAD, and MCP) increase with the number of edges in the graphs. 
However, for the proposed GTF model, the computational times of the spectral and SA algorithms are not sensitive to the edge size of the graph.

\begin{table*}[h]
\setlength\belowcaptionskip{1pt}
\caption{Misclassification rates averaged over 100 trials, with the variance below the rates.}
\label{tab:1}
\centering
\normalsize
\renewcommand{\arraystretch}{1.33}
\begin{tabular}{c|cccccccc}
\hline \hline
              & heart          & wine           & iris           & breast         & car            & wine-qual.     & ads            & yeast          \\ \hline 
\# of classes & 2              & 3              & 3              & 2              & 4              & 5              & 2              & 10             \\
\# of samples & 303            & 178            & 150            & 569            & 1,728          & 1,599          & 3,279          & 1,484          \\ \hline \hline
$\ell_1$            &\begin{tabular}[c]{@{}c@{}}0.158\\ (4e-4)\end{tabular}            &  \begin{tabular}[c]{@{}c@{}}0.040\\ (2e-4)\end{tabular}          &\begin{tabular}[c]{@{}c@{}}0.051\\ (5e-4)\end{tabular}            & \begin{tabular}[c]{@{}c@{}}0.058\\ (4e-4)\end{tabular}          & \begin{tabular}[c]{@{}c@{}}0.170\\ (6e-5)\end{tabular}           &\begin{tabular}[c]{@{}c@{}}0.574\\ (7e-4)\end{tabular}         &\begin{tabular}[c]{@{}c@{}}0.152\\ (1e-4)\end{tabular}         &\begin{tabular}[c]{@{}c@{}}0.391\\ (4e-4)\end{tabular}        \\\hline
SCAD          &\begin{tabular}[c]{@{}c@{}}0.150\\ (4e-2)\end{tabular}       &\begin{tabular}[c]{@{}c@{}}0.041\\ (5e-4)\end{tabular}           &\begin{tabular}[c]{@{}c@{}}0.049\\ (2e-4)\end{tabular}           &\begin{tabular}[c]{@{}c@{}}0.045\\ (2e-4)\end{tabular}           &\begin{tabular}[c]{@{}c@{}}0.149\\ (3e-5)\end{tabular}          &\begin{tabular}[c]{@{}c@{}}0.574\\ (1e-5)\end{tabular}           &\begin{tabular}[c]{@{}c@{}}0.153\\ (4e-4)\end{tabular}           &\begin{tabular}[c]{@{}c@{}}0.673\\ (3e-5)\end{tabular}           \\\hline
MCP           &\begin{tabular}[c]{@{}c@{}}0.148\\ (3e-4)\end{tabular}           &\begin{tabular}[c]{@{}c@{}}0.041\\ (1e-4)\end{tabular}            &\begin{tabular}[c]{@{}c@{}}0.048\\ (3e-4)\end{tabular}           &\begin{tabular}[c]{@{}c@{}}0.041\\ (2e-4)\end{tabular}           &\begin{tabular}[c]{@{}c@{}}0.148 \\ (1e-5)\end{tabular}          &\begin{tabular}[c]{@{}c@{}}0.572 \\ (1e-6)\end{tabular}          &\begin{tabular}[c]{@{}c@{}}0.150\\ (5e-4)\end{tabular}           &\begin{tabular}[c]{@{}c@{}}0.507 \\ (1e-5)\end{tabular}          \\\hline
$\ell_{2,0}$ (see \cref{sec_map})     & \textbf{\begin{tabular}[c]{@{}c@{}}0.143\\ (2e-4)\end{tabular}}& \textbf{\begin{tabular}[c]{@{}c@{}}0.035\\ (6e-5)\end{tabular}}  & \textbf{\begin{tabular}[c]{@{}c@{}}0.038\\ (3e-4)\end{tabular}} & \textbf{\begin{tabular}[c]{@{}c@{}}0.033\\ (1e-4)\end{tabular}} & \textbf{\begin{tabular}[c]{@{}c@{}}0.098\\ (1e-6)\end{tabular}} & \textbf{\begin{tabular}[c]{@{}c@{}}0.373\\ (3e-4)\end{tabular}} & \textbf{\begin{tabular}[c]{@{}c@{}}0.079\\ (5e-4)\end{tabular}} & \textbf{\begin{tabular}[c]{@{}c@{}}0.372\\ (5e-5)\end{tabular}}
\end{tabular}
\end{table*}
\raggedbottom

\subsection{Semi-supervised classification}
We now report results on the Modified Absorption Problem (MAP).
We compare the proposed MAP with the $\ell_{2,0}$-norm GTF regularization \eqref{eq:ssl} with other MAP-GTF models proposed in~\cite{pmlr-v38-wang15d,8926407}, which include MAP with $\ell_1$-norm GTF regularization~\cite{pmlr-v38-wang15d}, MAP with SCAD penalized GTF regularization~\cite{8926407}, and a MAP with MCP penalized GTF regularization~\cite{8926407}.

\subsubsection{Dataset}
We test the models on 8 popular classification datasets from \cite{UCIml}. 
For each dataset, we construct a $5$-nearest neighbor graph based on the Euclidean distance between provided features.
We randomly pick $20\%$ samples from the dataset and treat them as labeled samples.
For each dataset, we compute the misclassification rates over 100 repetitions.

\subsubsection{Parameter tuning}
For all the MAP-GTF models, we set $\epsilon=0.01$ (the regularization parameter in \eqref{eq:ssl}) and $R_{ij} = 1/K$.
For each model, we screen through the hyper-parameters and report the best results.

\subsubsection{The results}
Table~\ref{tab:1} summarizes the misclassification rate for all the methods over 100 repetitions of randomly selected labeled samples. 
The proposed MAP-GTF model \eqref{eq:ssl} achieves the smallest misclassification rate. 
We further do a t-test based on the misclassification rates over 100 repetitions and find that the misclassification rates of the proposed MAP-GTF model \eqref{eq:ssl} for all the datasets are significantly smaller than other methods ($p$-value is less than 0.001).

\subsubsection{Concluding remarks}
Lastly, we comment on the comparisons of these  GTF regularizers on MAP.
The experiment results all show that the $\ell_1$ model while being convex and having the least number of parameters (see Table~\ref{table:gtfmodels}), is consistently having the worst performance among all the tested models.
For the remaining nonconvex models, the $\ell_{2,0}$ model  consistently outperforms SCAD and MCP, and has fewer parameters to tune.
We believe that the superior performance of the MAP model with the $\ell_{2,0}$-norm GTF regularization in such a broad range of experiments is convincing evidence for the utility of the $\ell_{2,0}$-norm GTF regularizer.

\section{Conclusion}
We propose a first-order GTF model with $\ell_{2,0}$-norm penalty.
By converting it into a mixed integer program, we prove that the model is equivalent to a k-means clustering with graph cut regularization (Theorem~\ref{thm:1}). 
We propose two methods to solve the problem: approximate spectral method and simulated annealing. 
Empirically, we demonstrate the performance improvements of the proposed GTF model on both synthetic and real datasets for support recovery, signal estimation, and semi-supervised classification.

{As with many other graph signal processing models~\cite{7208894,7117446,7605501}, the model assumption of our proposed GTF model is that the graph structure should represent the data relationships. Therefore, the performance of our model heavily depends on how well the graph captures the structure of the data. Theorem~\ref{thm:1} provides insight into when the graph could help with the signal recovery. If the graph k-cut structure is consistent with k-means clustering of the signal,  the graph could help recover the inhomogeneous signal. When the graph is not available, we can follow the methods proposed in~\cite{8700659} to construct a graph for the proposed model.}

We believe that the superior performance of the proposed models in a broad range of experiments is convincing evidence for the utility of $\ell_{2,0}$-norm penalized GTF.
For future works, we think it will be interesting to investigate fast algorithms for solving the GTF model and study the error bound of the model. 
In addition, we believe that higher-order GTF models (see Fig.3 in \cite{pmlr-v38-wang15d}) with $\ell_0$-norm penalty also have great potential.

\section{Acknowledgement}

\appendices
\section{Proof of Theorem~\ref{thm:0}}~\label{append:0}
\begin{proof}
First we let $E = S_{=1} \cup S_{=0}$ and $S_{=1} \cap S_{=0} = \varnothing$, where
$$
S_{=0} \coloneqq  \Big\{ 
(i,j)\in E :
\mathbb{I}_{\neq 0} \big( \|\bm{b}_i-\bm{b}_j\|_2\big) = 0
\Big\}
.
$$
The proof has three parts: (i) show $S_{=1}$ is a graph cut, (ii) prove \eqref{eq:thm1_1} and (iii) show \eqref{eq:thm1_2}. 

\paragraph*{$S_{=1}$ is a graph cut}
We show that if we remove all edges in $S_{=1}$ from $G$, then for any edge $(i,j) \in S_{=1}$, the corresponding nodes $\{i, j \}$ are in different disjointed clusters.

\begin{itemize}
\item Assume after removing all edges in $S_{=1}$ from $G$ there exists an edge $(h,l)\in S_{=1}$ such that $\mathbb{I}_{\neq 0} \big( \|\bm{b}_h-\bm{b}_l\|_2\big)=1$ but the nodes $\{h, l\}$ are in the same partition.

\item As $G$ is connected, so nodes $\{h, l\}$ being in the same partition implies there exists a path $p =\{(h,s), \dots , (t, l)\}$ between $h$ and $l$. 

\item All the edges along the path $p$ are in $S_{=0}$ because by assumption $S_{=0}$ contains all edges after removing $S_{=1}$ from $G$. 
This implies 
$
\mathbb{I}_{\neq 0} \big( \|\bm{b}_h-\bm{b}_s\|_2\big) = \dots = 
\mathbb{I}_{\neq 0} \big( \|\bm{b}_t-\bm{b}_l\|_2\big) =0$ and thus $\mathbb{I}_{\neq 0} \big( \|\bm{b}_h-\bm{b}_l\|_2\big)=0$.
A contradiction to $\mathbb{I}_{\neq 0} \big( \|\bm{b}_h-\bm{b}_l\|_2\big)=1$.
\end{itemize}

\paragraph*{Prove \eqref{eq:thm1_1}}
$S_{=1}$ is a cut and it cuts $G$ into $k \in \IZ^+$ ($k\leq n$ is an unknown number) partitions.
Let $\bm{X} \in\{0,1\}^{n \times k}$ be the assignment matrix (where the $i$th row indicate the cluster membership of the $i$th row of $\bm{Y}$) that encodes the cut $S_{=1}$, and let $\bm{L}$ be the Laplacian of $G$,
we have the cut size 
\[
\begin{array}{rll}
&
~~ \dfrac{1}{2}\Tr(\bm{X}^\top \bm{L}\bm{X}) 
~=~
|S_{=1}| 
&
\text{see \cite{Fan2010,books/daglib/0004338}}
\vspace{-4mm}
\\
&=
\displaystyle
\sum_{(i,j)\in S_{=1}}  \mathbb{I}_{\neq 0} \big( \|\bm{b}_i-\bm{b}_j\|_2\big)
& \text{by definition of $S_{=1}$}
\vspace{-4mm}
\\
&= 
\displaystyle
\sum_{(i,j)\in S_{=1}}  \mathbb{I}_{\neq 0} \big( \|\bm{b}_i-\bm{b}_j\|_2\big)
\vspace{-4mm}
\\
&~~~~+ 
\underbrace{\sum_{(i,j)\in S_{=0}} \mathbb{I}_{\neq 0} \big( \|\bm{b}_i-\bm{b}_j\|_2\big)}_{= 0 }
& \text{by definition of $S_{=0}$}
\vspace{-4mm}
\\
&=
\displaystyle
\sum_{(i,j)\in E} \mathbb{I}_{\neq 0} \big( \|\bm{b}_i-\bm{b}_j\|_2\big)
.
\end{array}
\]

\paragraph*{Prove \eqref{eq:thm1_2}} 
After removing edges in the cut $S_{=1}$ from $G$, the graph $G$ will be separated into $k$ disjoint partitions.
Denote such partition as $C=\{C_1, ..., C_k\}$. 
\begin{itemize}
    \item For each partition,
    \begin{itemize}
        \item the nodes within the same partition are connected (because $G$ is connected); and
        \item all the edges connecting these nodes are in $S_{=0}$ (by $S_{=0} = E \,\backslash\, S_{=1}$).
    \end{itemize}
    
    \item By definition of $S_{=0}$, the signal of all the nodes in each partition share the same value, we denote such value by $\bm{\mu}_i$.
    
    \item As $\bm{\mu}_i$ represents the signal value for all the nodes in partition $C_i$, using the assignment matrix $\bm{X}$ gives \eqref{eq:thm1_2}.
\end{itemize}

\end{proof}

\bibliographystyle{IEEEtran}
  \bibliography{references}

% Generated by IEEEtran.bst, version: 1.14 (2015/08/26)
\begin{thebibliography}{10}
\providecommand{\url}[1]{#1}
\csname url@samestyle\endcsname
\providecommand{\newblock}{\relax}
\providecommand{\bibinfo}[2]{#2}
\providecommand{\BIBentrySTDinterwordspacing}{\spaceskip=0pt\relax}
\providecommand{\BIBentryALTinterwordstretchfactor}{4}
\providecommand{\BIBentryALTinterwordspacing}{\spaceskip=\fontdimen2\font plus
\BIBentryALTinterwordstretchfactor\fontdimen3\font minus
  \fontdimen4\font\relax}
\providecommand{\BIBforeignlanguage}[2]{{%
\expandafter\ifx\csname l@#1\endcsname\relax
\typeout{** WARNING: IEEEtran.bst: No hyphenation pattern has been}%
\typeout{** loaded for the language `#1'. Using the pattern for}%
\typeout{** the default language instead.}%
\else
\language=\csname l@#1\endcsname
\fi
#2}}
\providecommand{\BIBdecl}{\relax}
\BIBdecl

\bibitem{7208894}
S.~Chen, R.~Varma, A.~Sandryhaila, and J.~Kovačević, ``Discrete {S}ignal
  {P}rocessing on {G}raphs: {S}ampling {T}heory,'' \emph{IEEE Transactions on
  Signal Processing}, vol.~63, no.~24, pp. 6510--6523, 2015.

\bibitem{7117446}
S.~Chen, A.~Sandryhaila, J.~M.~F. Moura, and J.~Kovačević, ``Signal
  {R}ecovery on {G}raphs: {V}ariation {M}inimization,'' \emph{IEEE Transactions
  on Signal Processing}, vol.~63, no.~17, pp. 4609--4624, 2015.

\bibitem{7605501}
D.~Romero, M.~Ma, and G.~B. Giannakis, ``Kernel-{B}ased {R}econstruction of
  {G}raph {S}ignals,'' \emph{IEEE Transactions on Signal Processing}, vol.~65,
  no.~3, pp. 764--778, 2017.

\bibitem{4526700}
A.~Elmoataz, O.~Lezoray, and S.~Bougleux, ``Nonlocal {D}iscrete
  {R}egularization on {W}eighted {G}raphs: A {F}ramework for {I}mage and
  {M}anifold {P}rocessing,'' \emph{IEEE Transactions on Image Processing},
  vol.~17, no.~7, pp. 1047--1060, 2008.

\bibitem{6494675}
D.~I. Shuman, S.~K. Narang, P.~Frossard, A.~Ortega, and P.~Vandergheynst, ``The
  {E}merging {F}ield of {S}ignal {P}rocessing on {G}raphs: {E}xtending
  high-dimensional data analysis to networks and other irregular domains,''
  \emph{IEEE Signal Processing Magazine}, vol.~30, no.~3, pp. 83--98, 2013.

\bibitem{8347162}
A.~Ortega, P.~Frossard, J.~Kovačević, J.~M.~F. Moura, and P.~Vandergheynst,
  ``Graph {S}ignal {P}rocessing: {O}verview, {C}hallenges, and
  {A}pplications,'' \emph{Proceedings of the IEEE}, vol. 106, no.~5, pp.
  808--828, 2018.

\bibitem{10.1007/978-3-540-27819-1_43}
M.~Belkin, I.~Matveeva, and P.~Niyogi, ``Regularization and {S}emi-supervised
  {L}earning on {L}arge {G}raphs,'' in \emph{Learning Theory}, J.~Shawe-Taylor
  and Y.~Singer, Eds.\hskip 1em plus 0.5em minus 0.4em\relax Berlin,
  Heidelberg: Springer Berlin Heidelberg, 2004, pp. 624--638.

\bibitem{10.1007/978-3-540-72823-8_12}
S.~Bougleux, A.~Elmoataz, and M.~Melkemi, ``Discrete {R}egularization on
  {W}eighted {G}raphs for {I}mage and {M}esh {F}iltering,'' in \emph{Scale
  Space and Variational Methods in Computer Vision}, F.~Sgallari, A.~Murli, and
  N.~Paragios, Eds.\hskip 1em plus 0.5em minus 0.4em\relax Berlin, Heidelberg:
  Springer Berlin Heidelberg, 2007, pp. 128--139.

\bibitem{10.1109/FOCS.2007.66}
D.~A. Spielman, ``Spectral {G}raph {T}heory and its {A}pplications,'' in
  \emph{48th Annual IEEE Symposium on Foundations of Computer Science
  (FOCS'07)}.\hskip 1em plus 0.5em minus 0.4em\relax IEEE, 2007, pp. 29--38.

\bibitem{10.5555/3041838.3041953}
X.~Zhu, Z.~Ghahramani, and J.~Lafferty, ``Semi-{S}upervised {L}earning {U}sing
  {G}aussian {F}ields and {H}armonic {F}unctions,'' in \emph{Proceedings of the
  20th International conference on Machine learning (ICML-03)}, 2003, pp.
  912--919.

\bibitem{8926407}
R.~Varma, H.~Lee, J.~Kovačević, and Y.~Chi, ``Vector-{V}alued {G}raph {T}rend
  {F}iltering {W}ith {N}on-{C}onvex {P}enalties,'' \emph{IEEE Transactions on
  Signal and Information Processing over Networks}, vol.~6, pp. 48--62, 2020.

\bibitem{pmlr-v38-wang15d}
Y.~Wang, J.~Sharpnack, A.~J. Smola, and R.~J. Tibshirani, ``Trend {F}iltering
  on {G}raphs,'' \emph{Journal of Machine Learning Research}, vol.~17, no. 105,
  pp. 1--41, 2016.

\bibitem{pmid21527005}
G.~A. Pavlopoulos, M.~Secrier, C.~N. Moschopoulos, T.~G. Soldatos, S.~Kossida,
  J.~Aerts, R.~Schneider, and P.~G. Bagos, ``{{U}sing graph theory to analyze
  biological networks},'' \emph{BioData Mining}, vol.~4, p.~10, Apr 2011.

\bibitem{szklarczyk2023string}
D.~Szklarczyk, R.~Kirsch, M.~Koutrouli, K.~Nastou, F.~Mehryary, R.~Hachilif,
  A.~L. Gable, T.~Fang, N.~T. Doncheva, S.~Pyysalo \emph{et~al.}, ``The
  {STRING} database in 2023: protein--protein association networks and
  functional enrichment analyses for any sequenced genome of interest,''
  \emph{Nucleic {A}cids {R}esearch}, vol.~51, no.~D1, pp. D638--D646, 2023.

\bibitem{Kim2022-fy}
C.~Y. Kim, S.~Baek, J.~Cha, S.~Yang, E.~Kim, E.~M. Marcotte, T.~Hart, and
  I.~Lee, ``{HumanNet} v3: an improved database of human gene networks for
  disease research,'' \emph{Nucleic Acids Research}, vol.~50, no.~D1, pp.
  D632--D639, Jan. 2022.

\bibitem{Rodchenkov2020-fr}
I.~Rodchenkov, O.~Babur, A.~Luna, B.~A. Aksoy, J.~V. Wong, D.~Fong, M.~Franz,
  M.~C. Siper, M.~Cheung, M.~Wrana, H.~Mistry, L.~Mosier, J.~Dlin, Q.~Wen,
  C.~O'Callaghan, W.~Li, G.~Elder, P.~T. Smith, C.~Dallago, E.~Cerami,
  B.~Gross, U.~Dogrusoz, E.~Demir, G.~D. Bader, and C.~Sander, ``Pathway
  {C}ommons 2019 {U}pdate: integration, analysis and exploration of pathway
  data,'' \emph{Nucleic Acids Research}, vol.~48, no.~D1, pp. D489--D497, Jan.
  2020.

\bibitem{Hofree2013-ot}
M.~Hofree, J.~P. Shen, H.~Carter, A.~Gross, and T.~Ideker, ``Network-based
  stratification of tumor mutations,'' \emph{Nature Methods}, vol.~10, no.~11,
  pp. 1108--1115, Nov. 2013.

\bibitem{pmid12934013}
J.~M. Stuart, E.~Segal, D.~Koller, and S.~K. Kim, ``{{A} gene-coexpression
  network for global discovery of conserved genetic modules},'' \emph{Science},
  vol. 302, no. 5643, pp. 249--255, Oct 2003.

\bibitem{Chung:1997}
F.~R.~K. Chung, \emph{Spectral Graph Theory}.\hskip 1em plus 0.5em minus
  0.4em\relax American Mathematical Society, 1997.

\bibitem{10.1007/978-3-642-00202-1_24}
M.~Mahajan, P.~Nimbhorkar, and K.~Varadarajan, ``The planar k-means problem is
  {NP}-hard,'' in \emph{WALCOM: Algorithms and Computation}, S.~Das and
  R.~Uehara, Eds.\hskip 1em plus 0.5em minus 0.4em\relax Berlin, Heidelberg:
  Springer Berlin Heidelberg, 2009, pp. 274--285.

\bibitem{books/daglib/0004338}
V.~Vazirani, \emph{Approximation {A}lgorithms}.\hskip 1em plus 0.5em minus
  0.4em\relax Springer, 2001.

\bibitem{Newman2006}
M.~Newman, ``Finding community structure in networks using the eigenvectors of
  matrices,'' \emph{Physical Review E}, vol.~74, no.~3, Sep. 2006.

\bibitem{udell2019big}
M.~Udell and A.~Townsend, ``Why {A}re {B}ig {D}ata {M}atrices {A}pproximately
  {L}ow {R}ank?'' \emph{SIAM Journal on Mathematics of Data Science}, vol.~1,
  no.~1, pp. 144--160, 2019.

\bibitem{Alpert1999}
C.~Alpert, A.~Kahng, and S.-Z. Yao, ``Spectral partitioning with multiple
  eigenvectors,'' \emph{Discrete Applied Mathematics}, vol.~90, no. 1-3, pp.
  3--26, Jan. 1999.

\bibitem{Bornholdt2006}
J.~Reichardt and S.~Bornholdt, ``Statistical mechanics of community
  detection,'' \emph{Physical Review E}, vol.~74, p. 016110, Jul 2006.

\bibitem{Carlon2013}
E.~Carlon, ``{Computational Physics},'' KU Leuven, Tech. Rep., 2013.

\bibitem{10.2307/2334940}
W.~K. Hastings, ``Monte {C}arlo sampling methods using {M}arkov chains and
  their applications,'' \emph{Biometrika}, vol.~57, no.~1, pp. 97--109, 1970.

\bibitem{10.1093/oso/9780198803195.003.0009}
M.~P. Allen and D.~J. Tildesley, ``{Advanced Monte Carlo methods},'' in
  \emph{{Computer Simulation of Liquids}}.\hskip 1em plus 0.5em minus
  0.4em\relax Oxford University Press, 06 2017.

\bibitem{10.1093/oso/9780195099713.001.0001}
T.~Bäck, \emph{{Evolutionary Algorithms in Theory and Practice: Evolution
  Strategies, Evolutionary Programming, Genetic Algorithms}}.\hskip 1em plus
  0.5em minus 0.4em\relax Oxford University Press, 02 1996.

\bibitem{SELIM19911003}
S.~Z. Selim and K.~Alsultan, ``A simulated annealing algorithm for the
  clustering problem,'' \emph{Pattern Recognition}, vol.~24, no.~10, pp.
  1003--1008, 1991.

\bibitem{Johnson1989}
D.~S. Johnson, C.~R. Aragon, L.~A. McGeoch, and C.~Schevon, ``Optimization by
  {S}imulated {A}nnealing: {A}n {E}xperimental {E}valuation; {P}art {I},
  {G}raph {P}artitioning,'' \emph{Operations Research}, vol.~37, no.~6, pp.
  865--892, Dec. 1989.

\bibitem{Talukdar2009}
P.~Talukdar and K.~Crammer, ``New {R}egularized {A}lgorithms for {T}ransductive
  {L}earning,'' in \emph{Machine Learning and Knowledge Discovery in
  Databases}, 2009, pp. 442--457.

\bibitem{talukdar-pereira-2010-experiments}
P.~Talukdar and F.~Pereira, ``Experiments in {G}raph-based {S}emi-{S}upervised
  {L}earning {M}ethods for {C}lass-{I}nstance {A}cquisition,'' in
  \emph{Proceedings of the 48th Annual Meeting of the Association for
  Computational Linguistics}, Uppsala, Sweden, Jul. 2010, pp. 1473--1481.

\bibitem{https://doi.org/10.1002/nav.3800030109}
M.~Frank and P.~Wolfe, ``An algorithm for quadratic programming,'' \emph{Naval
  Research Logistics Quarterly}, vol.~3, no. 1-2, pp. 95--110, 1956.

\bibitem{10.5555/3042817.3042867}
M.~Jaggi, ``Revisiting {F}rank-{W}olfe: {P}rojection-{F}ree {S}parse {C}onvex
  {O}ptimization,'' in \emph{Proceedings of the 30th International Conference
  on Machine Learning, Proceedings of Machine Learning Research}, vol.~28, pp.
  17--19.

\bibitem{nr}
R.~A. Rossi and N.~K. Ahmed, ``The {N}etwork {D}ata {R}epository with
  {I}nteractive {G}raph {A}nalytics and {V}isualization,'' in \emph{Proceedings
  of the Twenty-Ninth AAAI Conference on Artificial Intelligence}.\hskip 1em
  plus 0.5em minus 0.4em\relax AAAI Press, 2015, p. 4292–4293.

\bibitem{HOLLAND1983109}
P.~W. Holland, K.~B. Laskey, and S.~Leinhardt, ``Stochastic blockmodels: First
  steps,'' \emph{Social Networks}, vol.~5, no.~2, pp. 109--137, 1983.

\bibitem{UCIml}
\BIBentryALTinterwordspacing
D.~Dua and C.~Graff, ``{UCI} {M}achine {L}earning {R}epository,'' 2017.
  [Online]. Available: \url{http://archive.ics.uci.edu/ml}
\BIBentrySTDinterwordspacing

\bibitem{8700659}
G.~Mateos, S.~Segarra, A.~G. Marques, and A.~Ribeiro, ``Connecting the dots:
  Identifying network structure via graph signal processing,'' \emph{IEEE
  Signal Processing Magazine}, vol.~36, no.~3, pp. 16--43, 2019.

\bibitem{Fan2010}
N.~Fan and P.~Pardalos, ``Multi-way clustering and biclustering by the {R}atio
  cut and {N}ormalized cut in graphs,'' \emph{Journal of Combinatorial
  Optimization}, vol.~23, no.~2, pp. 224--251, Sep. 2010.

\end{thebibliography}

\end{document}